\documentclass{article}
\usepackage{config/arxiv}
\usepackage[utf8]{inputenc} 
\usepackage[T1]{fontenc}    
\usepackage[colorlinks,citecolor=green,pagebackref,hypertexnames=false]{hyperref}    
\usepackage{url}            
\usepackage{booktabs}       
\usepackage{amsfonts}       
\usepackage{amsmath}
\usepackage{nicefrac}       
\usepackage{svg}
\usepackage{microtype}      
\usepackage{graphicx}
\usepackage{xcolor}
\usepackage{tablefootnote}
\usepackage{subcaption}
\usepackage{caption} 
\usepackage{setspace}

\usepackage{nicefrac}       
\usepackage[misc]{ifsym}
\usepackage{float}

\newcommand{\vx}{{\mathbf{x}}}
\newcommand{\vk}{{\mathbf{k}}}

\newcommand{\vy}{{\mathbf{y}}}

\newcommand{\vz}{{\mathbf{z}}}

\newcommand{\vc}{\mathbf{c}}

\newcommand{\fref}[1]{Fig.~\ref{fig:#1}}

\newcommand{\tref}[1]{Table~\ref{tab:#1}}

\def\figref#1#2{\includegraphics[width=#1\columnwidth]{figs/#2}} 

\title{Navigating Efficiency of MobileViT through Gaussian Process on Global Architecture Factors}

\author{
     Ke Meng \\
	School of Mathematics and Statistics, Central South University (CSU) \\
 	\texttt{Mengke2017@csu.edu.cn} \\
	\And	
    Kai Chen (\Letter) \\
	School of Mathematics and Statistics, Central South University (CSU) \\
	\texttt{kaichen6@csu.edu.cn}
}

\date{%
	$^1$Central South University (CSU)\\%
	\today}

\def\figref#1#2{\includegraphics[width=#1\columnwidth]{figure/#2}}  

\begin{document}
 
\maketitle

\begin{abstract}

Numerous techniques have been meticulously designed to achieve optimal architectures for convolutional neural networks (CNNs), yet a comparable focus on vision transformers (ViTs) has been somewhat lacking. Despite the remarkable success of ViTs in various vision tasks, their heavyweight nature presents challenges of computational costs. In this paper, we leverage the Gaussian process to systematically explore the nonlinear and uncertain relationship between performance and global architecture factors of MobileViT, such as resolution, width, and depth including the depth of inverted residual blocks and the depth of ViT blocks, and joint factors including resolution-depth and resolution-width. We present design principles twisting magic 4D cube of the global architecture factors that minimize model sizes and computational costs with higher model accuracy. 
We introduce a formula for downsizing architectures by iteratively deriving smaller MobileViT V2, all while adhering to a specified constraint of multiply-accumulate operations (MACs). 
Experiment results show that our formula significantly outperforms CNNs and mobile ViTs across diversified datasets.
\end{abstract}

\section{Introduction}
Light-weight convolutional neural networks (CNNs)\cite{ma2018shufflenet,howard2017mobilenets,sandler2018mobilenetv2,howard2019searching} have played a significant role in enabling various computer vision tasks on mobile devices. However, vision transformers (ViTs) are yet to find popular use on such devices. Unlike lightweight CNNs, which are relatively easy to optimize and seamlessly integrate with task-specific networks, ViTs\cite{dosovitskiy2020image,touvron2021training,xiao2021early,wang2021pyramid} are considerably heavy-weight in terms of computational cost and memory footprint. This makes their deployment and utilization of resource-constrained mobile devices a challenging endeavor.

Due to the exceptional performance of transformers across a range of tasks\cite{touvron2021training,chen2022mobile,bao2022vlmo,radford2021learning,devlin2018bert}, with a particular emphasis on the outstanding capabilities of vision transformers, an increasing number of applications and companies are keen to deploy transformers on edge devices. Popular transformers on edge devices include MobileViT\cite{mehta2021mobilevit} and MobileViT V2\cite{mehta2022separable} proposed by Apple and Mobile-former\cite{chen2022mobile} proposed by Microsoft etc. Expanding on MobileViT\cite{mehta2021mobilevit}, our goal is to thoroughly explore how architecture factors including resolution, width, the depth of inverted residual blocks, the depth of ViT blocks, and joint factors including resolution-depth and resolution-width affect model's performance. These architecture factors determine the size, computational cost, and memory usage of MobileViT. Especially, to deploy the transformers on mobile devices, we continuously adjusted the architecture factors above to reduce memory cost and latency.
When given an upper limit of multiply-accumulate operations (MACs), we propose a valuable formula to recommend the ideal resolution and depth for MobileViT. The formula involves Gaussian process (GP) nonlinear fitting on the architecture factors of cutting-edge models\cite{chen2023compressible,chen2020incorporating}.

Specifically, inspired by the work \cite{han2020model}, we leverage GPs to fit the nonlinear and uncertain relationships between resolution, depth, width, and model performance. Interpreting the fitting of GP, we delve into twisting architecture factors to achieve an efficient and powerful MobileViT model. We empirically assess the impact of architecture factors and reveal that other rule \cite{tan2019efficientnet} loses its effectiveness when applied to MobileViT. 
Notably, for MobileViT architecture, we confirm that resolution and width are more important than depth.
Furthermore, we compare the improved MobileViT with well-known mobile-oriented models such as the MobileNet\cite{howard2017mobilenets}, and MobileNetV2\cite{sandler2018mobilenetv2}.
We outline our challenge, motivation, and contribution as follows.

\textbf{Challenge and motivation:} 
The ViT models are usually over-parameterized with hundreds of millions of parameters due to the use of heavy-weight attention structures, which are sometimes unusable for cost-sensitive applications. 
Despite the recent progress of ViT, current MobileViT models are still limited in their compressibility and computational complexity to perform vision tasks in mobile devices efficiently.
Current approaches to MobileViT do not provide insight into the nonlinear relationship between model performance and global architecture factors determining model complexity. The potential improvement of MobileViT can be informed by how the global architecture factors impact model performance. 

\textbf{Contribution:}
In response to the challenges faced, we leverage GP to reconstruct the global architecture of MobileViT. The key idea is to learn the importance of global architecture factors and control the compression rule of architecture factors. To the best of our knowledge, this is the first work investigating how twisting global architecture factors of MobileViT to achieve better performance and efficiency. Specifically, our contributions are as follows:
\begin{itemize}
\item We explore the nonlinear impact of a single global architecture factor in determining the performance and compressibility of MobileViT.

\item We delve into how the depth of MobileViT in terms of the depths of attention block and other blocks impacts the performance of MobileViT differently.

\item We provide insight into the nonlinear relationship between the performance of MobileViT and joint factors including resolution-depth and resolution-width by using a 2D grid GP.

\item To improve the extensive deployment of MobileViT, we propose a useful formula with the target MACs to squeeze the latent compression space of MobileViT and guide its global architecture adjustment.

\item We provide diversified experiments to construct computational MobileViT and draw a picture of how MobileViT performs given different global architecture factors.
\end{itemize}

\section{Transformer for vision task}
The transformer model was originally proposed by Google in 2017\cite{vaswani2017attention}, aiming to solve problems such as poor model performance and difficulty in training for long sequence inputs. Compared with traditional sequence models such as recurrent neural networks(RNNs) and CNNs, a transformer model has the following characteristics.
It does not need to maintain the order of time steps and can parallel process the entire input sequence, thus enabling the training of deeper models in a shorter time. Besides, it uses a self-attention mechanism, which can better capture long-term dependencies in the sequence. 

\textbf{Vision transformer:}
Besides, ViT is a variant of the transformer model that has been adapted for use in computer vision tasks, such as image classification, object detection, and semantic segmentation. ViT was first introduced by Google researchers in 2020 \cite{wu2020visual} and has gained popularity in the computer vision community due to its strong performance on a range of vision benchmarks.
The basic idea behind ViT is to apply the transformer model to image patches extracted from an input image, rather than to the entire image. The input image is divided into a grid of non-overlapping patches, which are then linearly projected to a lower-dimensional embedding space\cite{dosovitskiy2020image,wang2021pyramid}. The resulting embeddings are fed into the transformer model, which processes them using self-attention and multi-layer perceptron (MLP) modules to produce a final output.
One of the key capacities of ViT is that it can process variable-sized inputs\cite{dosovitskiy2020image}, unlike many CNN models that require fixed-size inputs. This allows ViT to handle images of different sizes without the need for resizing or cropping. Additionally, ViT has demonstrated strong performance on several image classification benchmarks, often outperforming state-of-the-art CNN models\cite{touvron2021training}.

\begin{figure}[h!]
\centering
\renewcommand{\tabcolsep}{2.0mm}
\begin{tabular}{p{0.5mm}*{1}{c}}
& \figref{0.7}{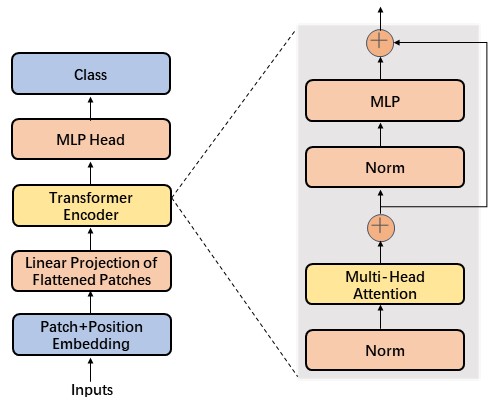} \\
\end{tabular}
\caption{A ViT model comprises an input layer, position and patch embeddings, a linear projection layer, and a self-attention module. The self-attention module consists of a normalization layer, a multi-head attention layer, and a fully connected layer. The black arrows represent the connections and data flow between these layers.}
\label{fig:vit_structure}
\end{figure}


\textbf{MobileViT:}
There are several excellent ViTs on mobile devices have been proposed. One such example is the MobileViT\cite{mehta2021mobilevit} introduced by Apple. MobileViT utilized the inverted residual block and MobileViT block to reduce the number of parameters. It is more suitable for deployment on mobile devices.
MobileViT is specifically designed for vision tasks on mobile devices. It replaces the traditional CNN with a self-attention module for image feature extraction and attention computation. This replacement allows the model to learn global image representations without being limited by fixed-size receptive fields.
The first version of MobileViT (denoted by MobileViT V1) is designed for 
accommodating resource constraints of mobile devices. It utilizes both depthwise separable convolutions and pointwise convolutions to reduce the number of parameters and computations of ViT. Additionally, MobileViT V1 employs model lightweight techniques such as channel attention mechanisms and width polynomial approximation to further reduce model complexity. The goal of MobileViT V1 is to maintain high performance while minimizing model size and computational costs, seeking real-time vision tasks on mobile devices.

\begin{figure}[h!]
\centering
\renewcommand{\tabcolsep}{2.0mm}
\begin{tabular}{p{0.5mm}*{1}{c}}
& \figref{0.7}{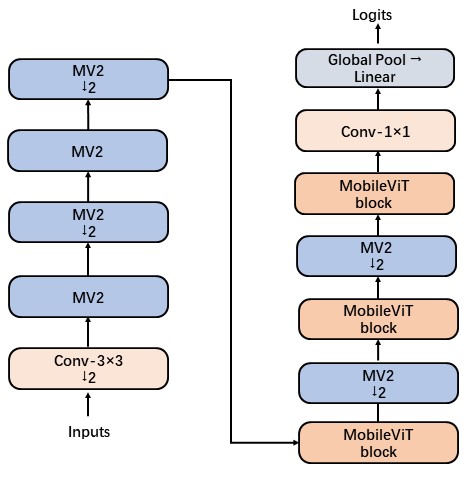} \\
\end{tabular}
\caption{The architecture of MobileViT. Notations Conv-$n\times n$ and MV2 denote standard $n\times n$ convolution and MobileNetV2 block respectively. Module with downsampling is indicated by $\downarrow 2$. The input feature goes through a convolutional layer that increases the channel dimension to a larger bottleneck channel size. 
}
\label{fig:mvit_structure}
\end{figure}


\textbf{MobileViT V2:} Replacing the multi-head attention (MHA) mechanism in MobileViT V1 with separable self-attention, we can obtain MobileViT V2. To achieve efficient inference, the separable self-attention method performs element-wise operations for the MHA mechanism. This replacement reduces the computational complexity and improves the overall efficiency of the original MobileViT during inference. By using element-wise operations, the model can perform attention calculations more efficiently while maintaining the ability to capture important patterns within the input data.

\textbf{Inverted residual block:} The design concept of inverted residuals aims to increase the model's representational capacity and maintain low computational complexity by modifying the traditional residual connection structure. The inverted residual block utilizes lightweight depthwise separable convolutions to adjust the input and output channels for feature map matching.
This block is widely used for lightweight deep models on mobile and embedded devices.

\textbf{MobileViT block:} To maintain an overall lightness of ViT and balance the usage of convolutions and transformers, MobileViT block replaces the local processing typically found in convolutions with global processing using transformers. This unique approach enables the MobileViT block to exhibit properties similar to both CNNs and ViTs. Therefore, the MobileViT block can learn more effective representations while utilizing fewer parameters and employing simpler training techniques.

\textbf{Separable self-attention}
Inspired by MHA, self-attention in traditional transformer models is computed for all token pairs in the sequence. This leads to a computational complexity of $\mathcal{O}(l^2)$, where $l$ is the sequence length.
Separable self-attention only computes the context score for a latent token $L$, which significantly reduces the computational cost to $\mathcal{O}(l)$\cite{mehta2022separable}. 

The input $\vx$ is processed
using three branches, i.e., input $\mathcal I$, key $\mathcal K$, and value $\mathcal V$. 
Specifically, 
the input $\vx$ is first linearly projected to a $d$-dimensional space using a key branch $\mathcal K$. The key branch $\mathcal K$ has weights $W_Q$ 
and produces an output $\vx_k$. 
The context vector $\vc_v$
is then calculated as a weighted sum of the projected tokens $\vx_k$. We use context scores $\vc_s$ to weight the projected tokens. 
The final context vector $\vc_v$ is computed efficiently and has the form as follows:
\begin{align}\label{eq:cv}
\vc_v=\sum\limits_{i=1}^l \vc_s(i)\vx_k(i).
\end{align}
The contextual information $\vc_v$ is shared across all tokens in $\vx$. The input $\vx$ is linearly transformed using a value branch $\mathcal{V}$ with weights ${W_V}$, followed by a ReLU activation, resulting in $\vx_v$. Contextual information is then incorporated into $\vx_v$ through element-wise multiplication. The final attention $\vz_{a}$ is obtained by passing this result through another linear layer with weights ${W_L}$.
Mathematically, separable self-attention is defined as:
\begin{align}\label{eq:sep-self-atten}
\vz_{a}=\left(\underbrace{\sum(\overbrace{\sigma(\vx {W_I})}^{\vc_s\in\mathbb{R}^k}* \vx{W_Q})}_{\vc_v\in\mathbb{R}^d}*\text{ReLU}(\vx {W_V})\right){W_L},
\end{align}
where * and $\sum$ are broadcastable element-wise multiplication and summation operations, respectively.

\begin{figure}[h!]
\centering
\renewcommand{\tabcolsep}{2.0mm}
\begin{tabular}{p{0.5mm}*{1}{c}}
& \figref{0.5}{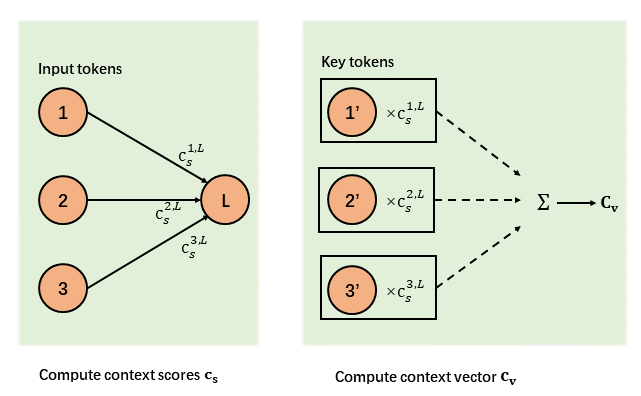} \\
\end{tabular}
\caption{Diagram of separable self-attention. The resultant vector is normalized using softmax to produce context scores $\vc_s$.
These context scores are used to weight key tokens and produce a context vector $\vc_v$, which encodes contextual information.}
\label{fig:separable-self-attn}
\end{figure}


\section{Related Work}\label{sec:rw}
\textbf{Advances for efficient ViTs:}
Advanced methods that improve the performance and reduce the complexity of transformers include mixed-precision training\cite{micikevicius2017mixed}, efficient optimizers\cite{dettmers20218}, and knowledge distillation\cite{touvron2021training}. Compression of transformers typically involves reducing the size of the model while preserving its accuracy. This can be done using various techniques used
in CNNs.


\textbf{Pruning:}
Removing unnecessary weights or connections of transformer \cite{han2015deep,li2016pruning,lin2020hrank,lin2019towards} is known as pruning. By identifying and eliminating the connections that contribute the least to the model's performance, pruning reduces the model's size and speeds up its operations without significantly sacrificing accuracy.

\textbf{Quantization:} \cite{gu2019projection,han2020training,courbariaux2016binarized,jacob2018quantization,rastegari2016xnor} Reducing the precision of weights and activation in a model refers to quantization. Quantization uses fewer bits to represent each weight or activation, such as using $8$-bit integer values instead of $32$-bit floating-point values. Quantization obtains a smaller model that performs inference faster, although there might be a slight decrease in accuracy.

\textbf{Distillation:}\cite{hinton2015distilling,romero2014fitnets,xu2019positive,you2017learning} In the original motivation of distillation, a smaller model can be trained to imitate the behavior of a larger, more complex model. During training, the outputs of the larger model serve as targets for the smaller model. By doing so, the smaller model can learn to replicate the performance of the larger model while being significantly smaller and faster.

\textbf{Activation sparsity:} To induce highly sparse activation maps without accuracy loss, kurtz\cite{kurtz2020inducing} introduces a new regularization technique, coupled with a new
threshold-based sparsification method based on a
parameterized activation function.
These techniques can be used individually or in combination to achieve the desired level of compression while maintaining acceptable levels of accuracy.

\textbf{Exponential moving average (EMA):}
EMA is a method used to stabilize and smooth the training process of deep neural networks (DNNs) \cite{howard2019searching,xiao2021early}. It involves maintaining a running average of model parameters throughout training. The EMA assigns exponentially decaying weights to the previous parameter values, with a higher weight given to more recent values. By incorporating this moving average into the training process, the model can benefit from the collective knowledge gained from earlier stages of training. This helps to reduce the impact of noisy or fluctuating gradients and improve the stability and generalization of the model. Like \cite{mehta2021mobilevit,mehta2022separable}, our work also employs EMA.

\textbf{Neural architecture search (NAS):} 
NAS automates the design of DNNs and contributes significantly to advancements in tasks such as image classification\cite{zoph2018learning,liu2017hierarchical}, object detection\cite{ghiasi2019fpn}, and semantic segmentation\cite{zhang2019customizable}. While traditional NAS methods rely on reinforcement learning (RL) \cite{zoph2018learning} and evolutionary algorithm (EA)\cite{real2019regularized}, they tend to be computationally intensive due to the huge search space. 
However, there are three main differences between NAS and our method. Firstly, the NAS is more suitable for the design of complex DNNs other than efficient MobileViT. Secondly, NAS primarily focuses on the adjustment of local structures in DNNs. The uncertainty of the adjustment is very large due to the various size and diversity of local neural network structures. Our method adjusts the global architecture of MobileViT in terms of resolution, width, and depths, which is more succinct and efficient.
Moreover, NAS typically explores an undetermined architecture of DNNs for each specific task\cite{li2020gp}. 
Our method employs the base architecture of MobileViT, which is determined.
Therefore, compared with the substantial computational cost of NAS, we pursue an efficient method for optimizing the global architecture factors and achieving high performance of MobileViT with minimal adjustments required.

\section{Rule and navigation of efficient MobileViT}\label{sec:approach}
In this section, we 
rethink the impact of architecture factors $(r, d_i, d_m, w)$ in different inputs' sizes, MobileViT networks, and compression rules, where 
$r$, $d_{i}$, $d_{m}$, and $w$ denote the resolution, depth of inverted residual block, the depth of MobileViT block, and width respectively. We introduce a novel concise learning formula twisting magic 4D cube
of the global architecture factors to produce efficient MobileViT.

\subsection{Key global architecture factors of MobileViT}\label{rdw}

\textbf{Resolution:} For ViTs, the resolution typically refers to the size of an input image. Through calculating, we can get the sequence length or the number of tokens of this image.  The resolution affects the memory and computational requirements of the model, as well as the ability to capture key information of an image.

\textbf{Factor of depth:} It refers to the number of attention layers in transformer models. Each layer consists of multiple self-attention heads and feed-forward neural networks. Increasing the depth of the model allows it to capture more complex patterns and dependencies in the visual data. However, a deeper model may require more computational resources. 
Training a deeper transformer without proper regularization techniques is challenging.

\textbf{Factor of width:} Width in ViTs indicates the dimensionality of token embeddings or the number of channels in the hidden layers. 
It determines the capacity or representational power of the model. 
A wider model can capture more intricate features and express complex patterns in the visual data. However, increasing the width also increases the number of parameters and computational requirements. We adjust the width by controlling the width multiplier in MobileViT V2 \cite{mehta2022separable}.

\textbf{Combination of architecture factors:}
Finding an appropriate balance between different factors in MobileViT is crucial for achieving good performance and efficiency on vision tasks. 
The optimal combination of these architecture factors is crucial for achieving good performance and efficiency on vision tasks, which depends on the specific dataset, computational resources, and task complexity. It often requires extraordinary and numerous experiments and fine-tuning to determine the appropriate architecture for a given task.

\subsection{Impact of single factor}
It is natural to investigate whether deeper or wider networks have a significant impact on MobileViT performance and computation complexity. Single factors, for instance, depth may play a role\cite{kurtz2020inducing} for the balance between model performance and complexity. 
Due to the powerful representation ability of ViT and the limitations of computing capacity in mobile devices, it is necessary to quantify the importance of resolution, depth, and width in ViTs, particularly in the case of MobileViTs. Recognizing the architecture of MobileViT, we propose to investigate the impact of architecture factors $(r, d_i, d_m, w)$ while keeping certain constraints fixed, such as MACs.
MACs represent the total number of multiplication and accumulation operations performed in a deep learning model. 
We use MACs to quantify the computational complexity of a MobileViT model and compare the computational efficiency of different MobileViT models.

We use a modeling problem concerning the accuracy of MobileViT to demonstrate how the model performance can be used to set global architecture factors in MobileViT. A GP is trained on our experimental evidence by maximizing the marginal likelihood (or evidence). The marginal likelihood refers to the marginalization over the latent function. The resulting GP model provides an excellent fit to the data as well as insights into its properties by interpretation of the posterior distribution. 
For the impact of single factors, the data is one-dimensional, and therefore the posterior distribution is easy to visualize. 
A total of four architecture factors are considered, which in practice rules out the use of cross-validation and NAS for setting architecture factors. 
Specifically, we denote the MACs of the given baseline MobileViT V2 as $m_0$. The resolution of an input image is denoted as $r_0\times r_{0}$, the width as $w_0$, and the depth as $d_0$. 

\begin{figure}[htb!]
\centering
\renewcommand{\tabcolsep}{-0.1mm}
\scriptsize
\begin{tabular}{p{0.5mm}*{2}{c}}
& \figref{0.35}{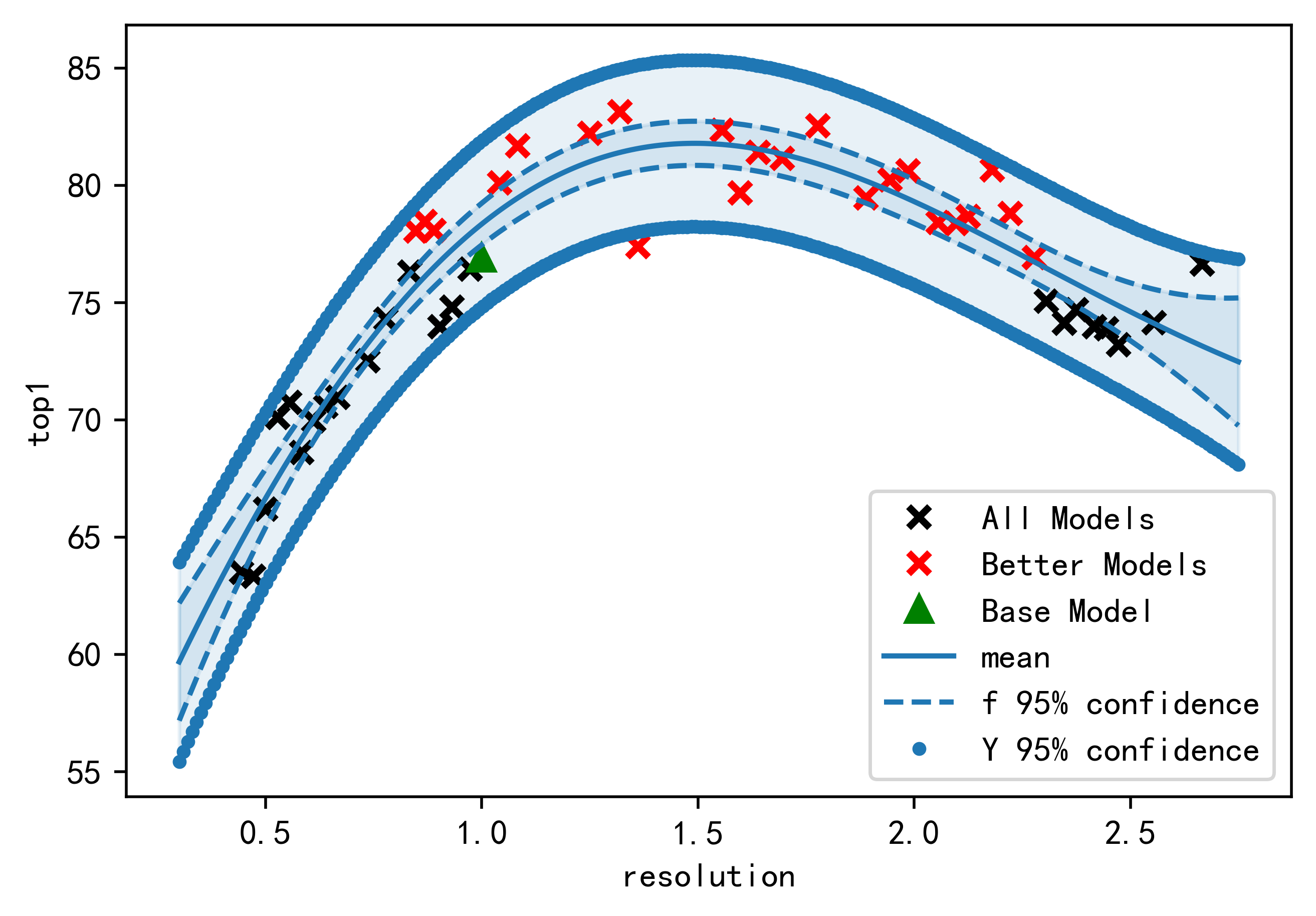} 
& \figref{0.35}{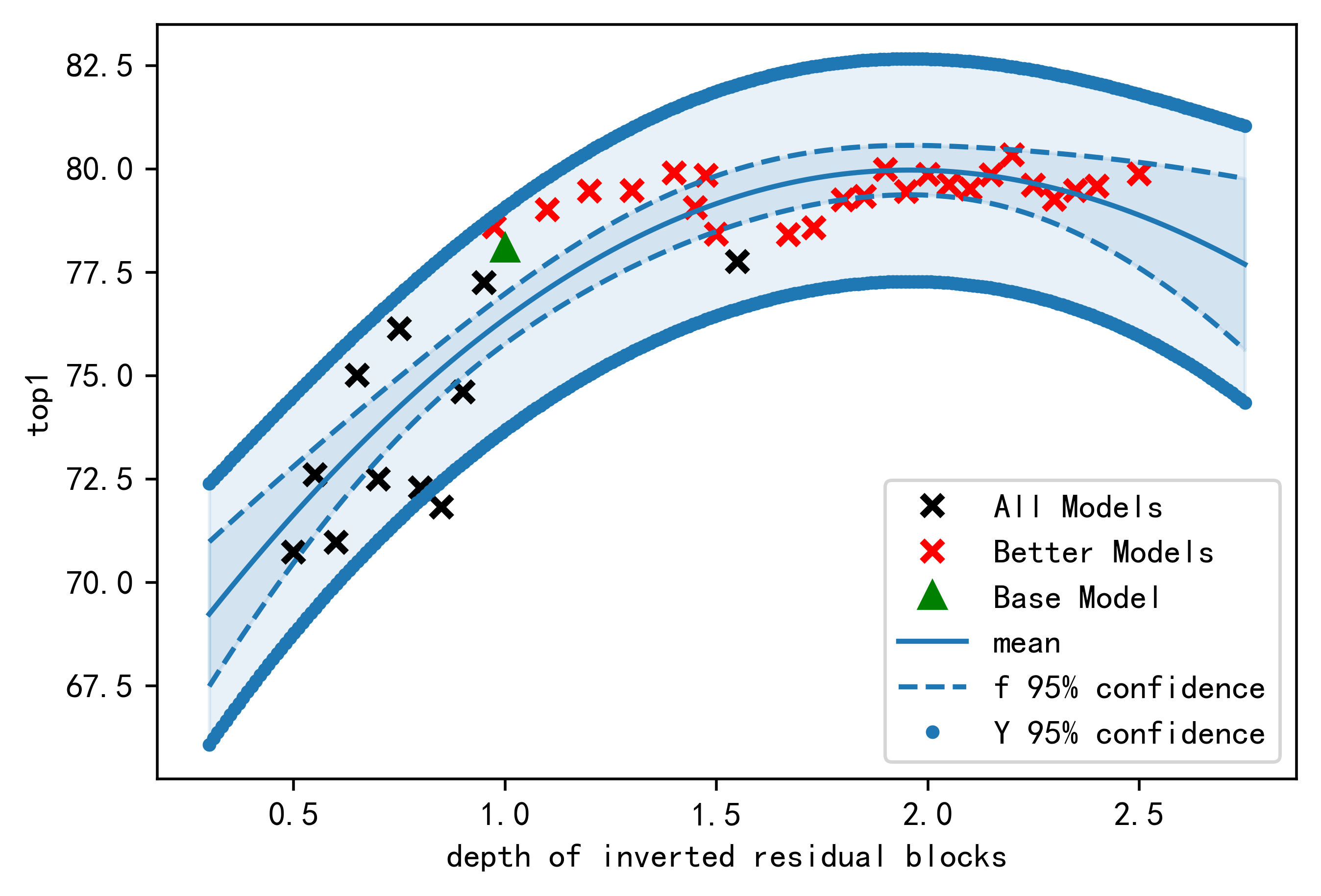} \\ 
& (a) Top 1 $\textit{v.s.}$ $r$ & (b) Top 1 $\textit{v.s.}$ $d_i$ \\
& \figref{0.35}{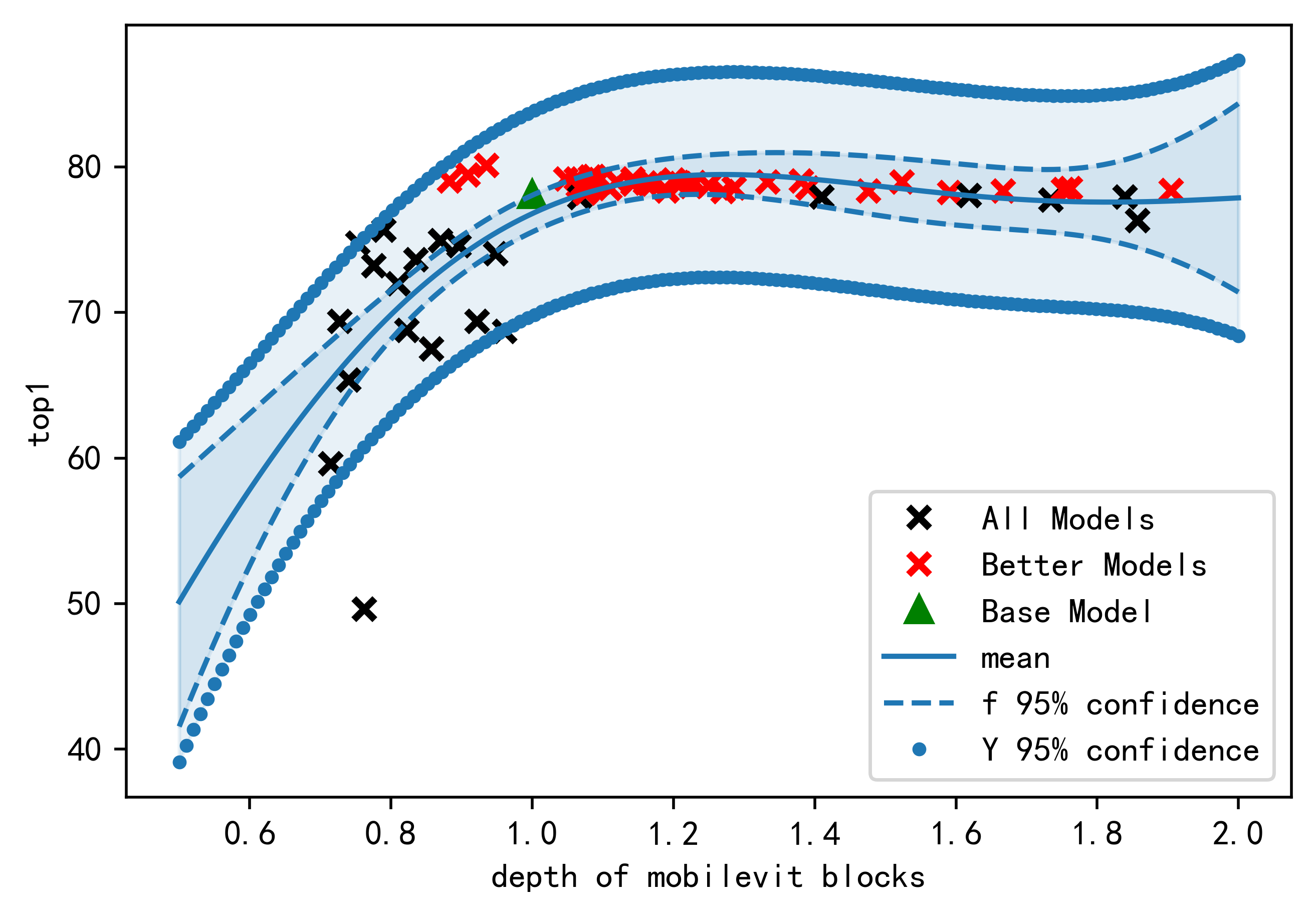}
& \figref{0.35}{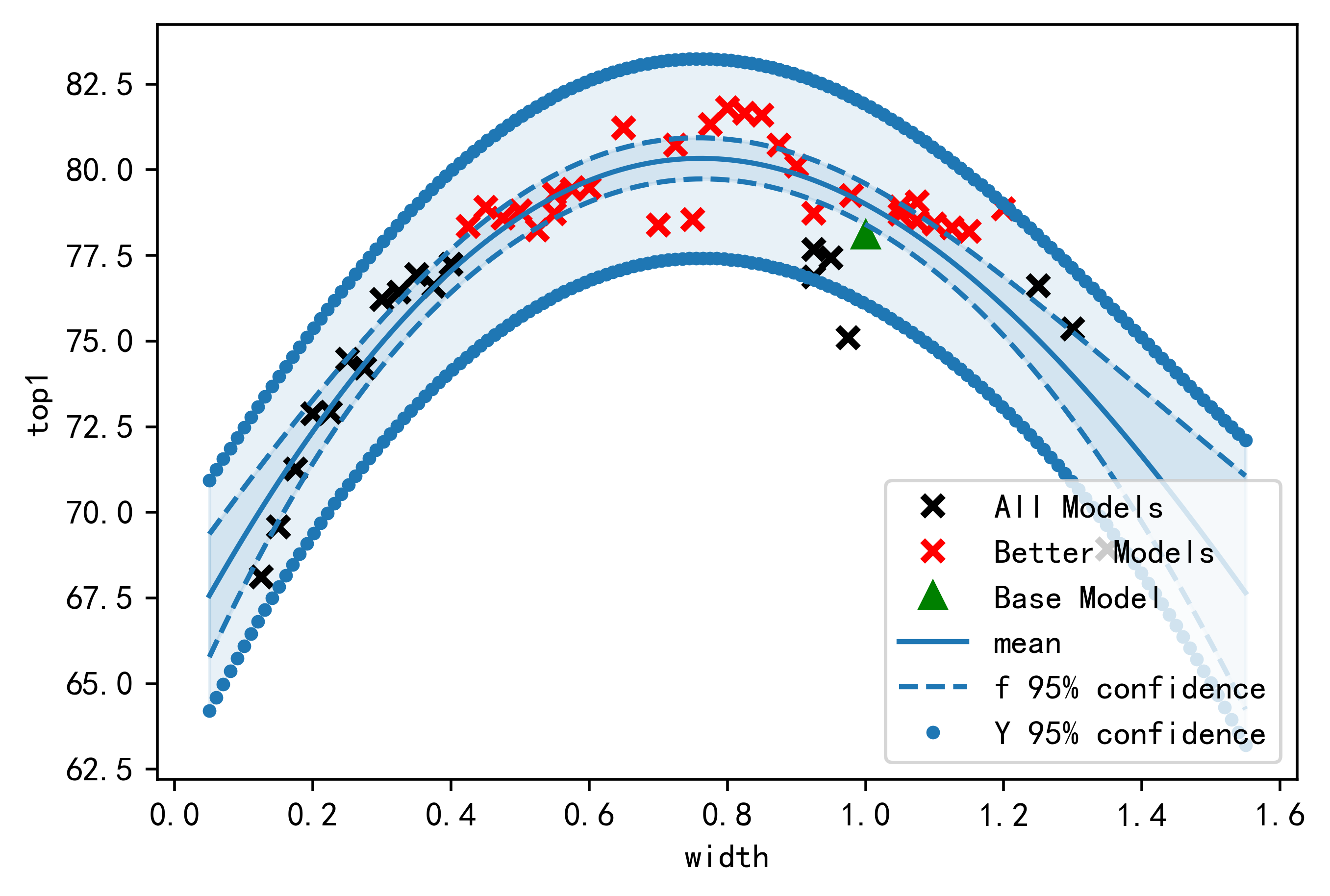} \\
& (c) Top 1 $\textit{v.s.}$ $d_m$ & (d) Top 1  $\textit{v.s.}$ $w$
\end{tabular}
\caption{Fitted top 1 accuracy $\textit{v.s.} (r, d_i, d_m, w)$ for MobileViT V2 models with $\sim$ 2000M MACs on ImageNet-100. There are 60 observations of top 1 accuracy, together with a 95\% predictive confidence interval (CI) for the GP regression model. Note also that the CI gets wider the further the predictions are extrapolated. The top 1 accuracy of the baseline MobileViT V2 model is labeled with a green triangle. Red crosses and black crosses denote the better models with higher accuracy than the baseline model and the inferior models with lower accuracy than the baseline, respectively. }
\label{fig:top1-rdw}
\end{figure}

\begin{figure}[h!]
\centering
\renewcommand{\tabcolsep}{-0.1mm}
\scriptsize
\begin{tabular}{p{0.5mm}*{2}{c}}
& \figref{0.35}{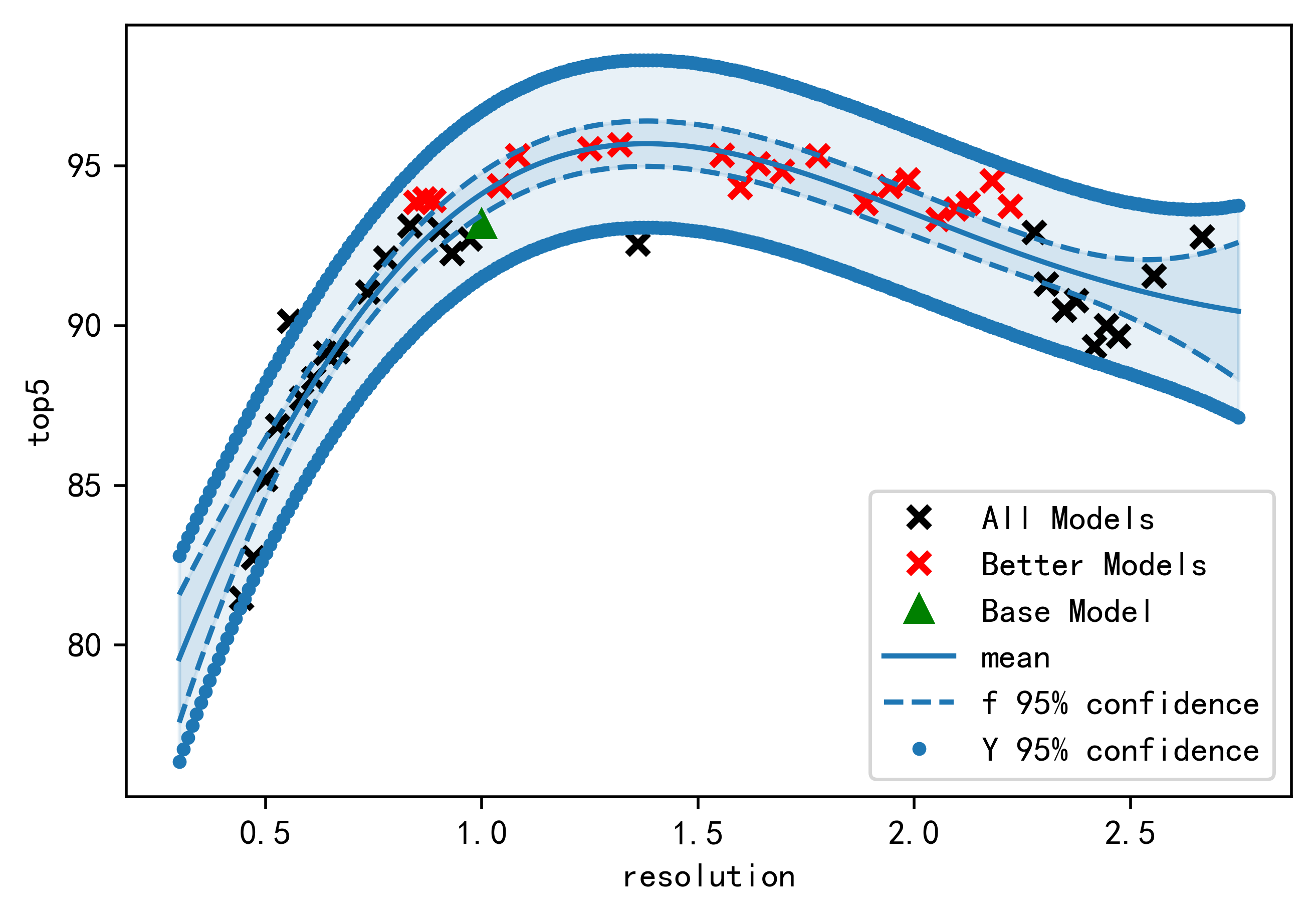} 
&  \figref{0.35}{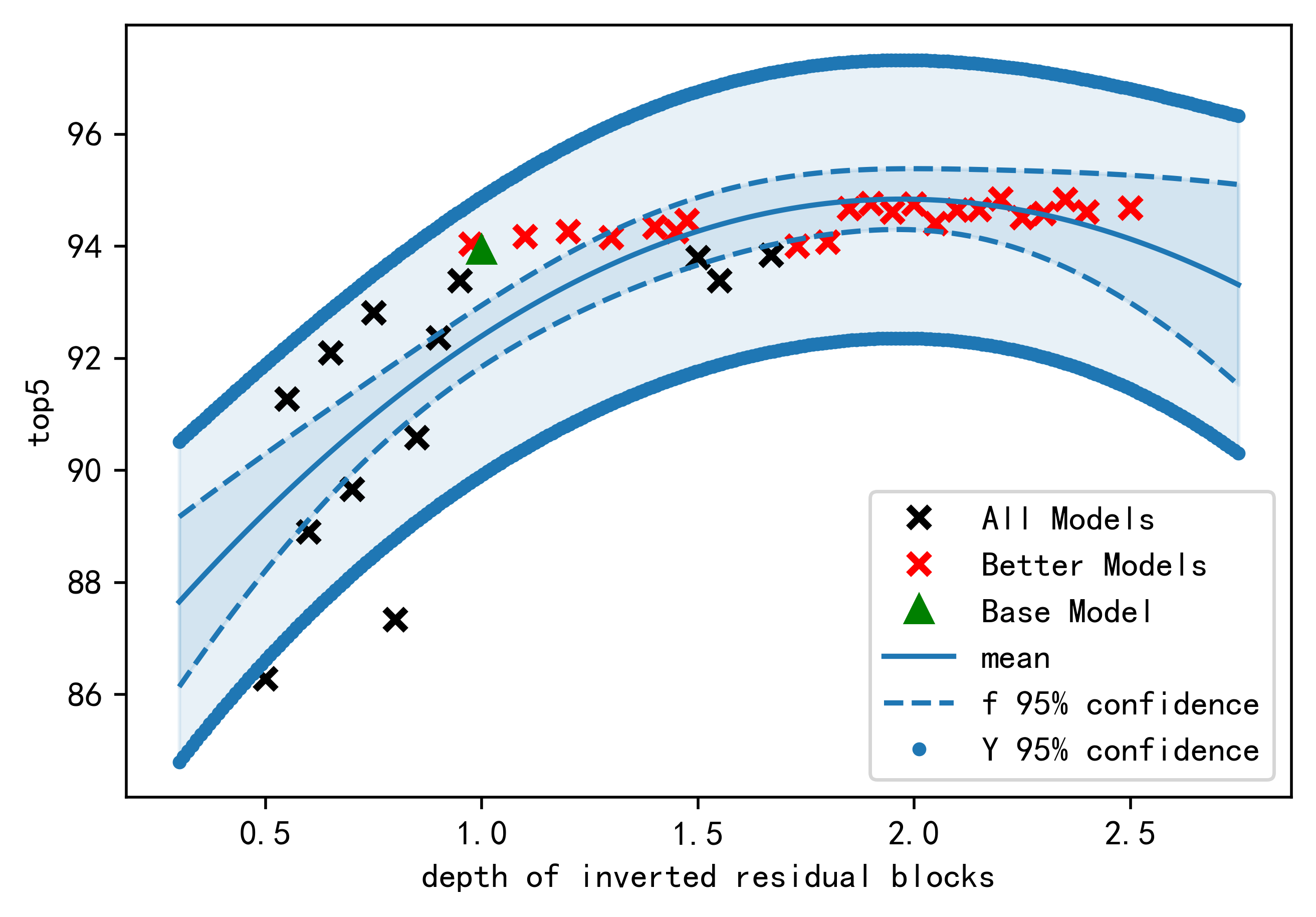} 
\\ 
& (a) Top 5 $\textit{v.s.}$ $r$ & (b) Top 5 $\textit{v.s.}$ $d_i$ \\
& \figref{0.35}{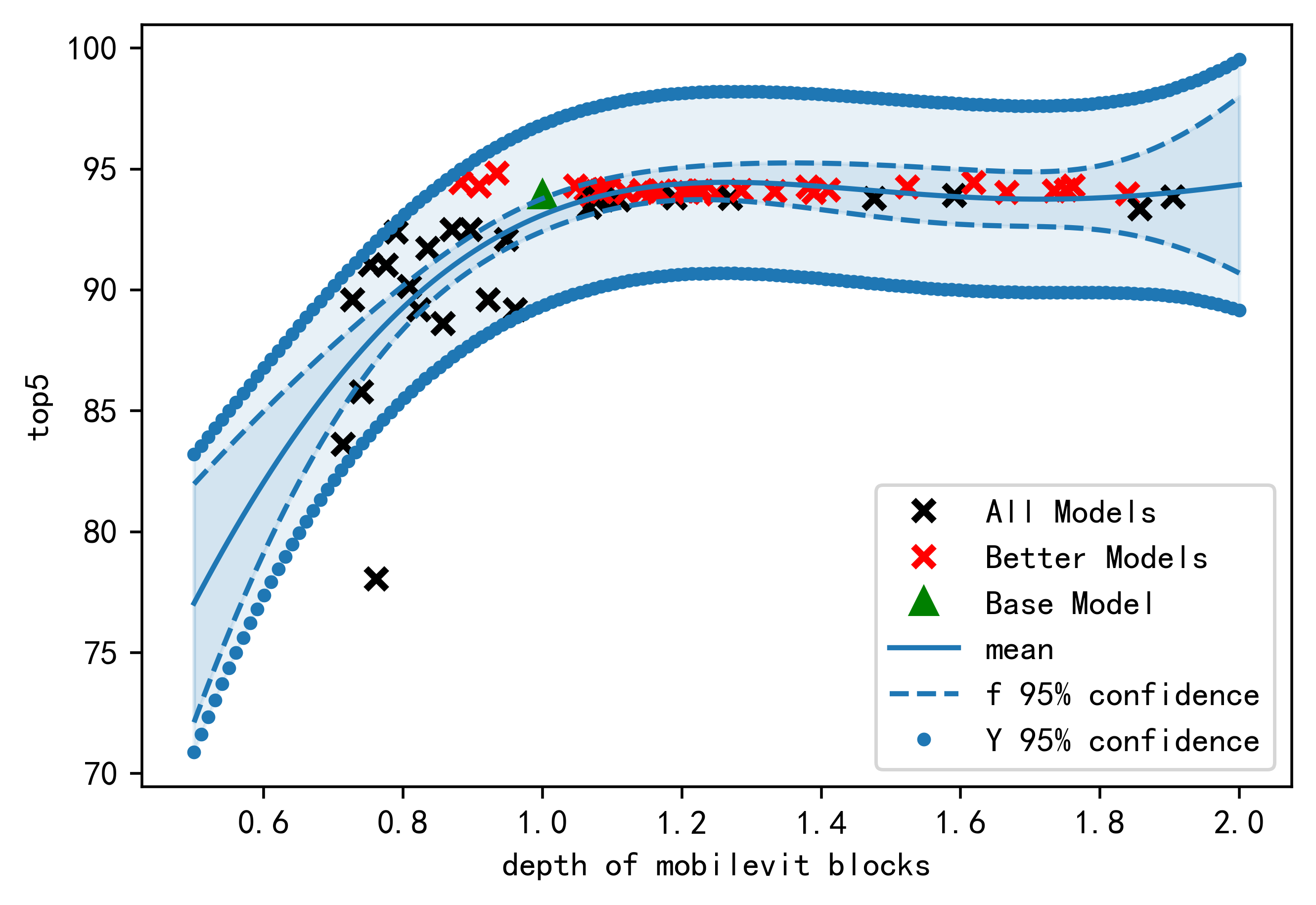}
& \figref{0.35}{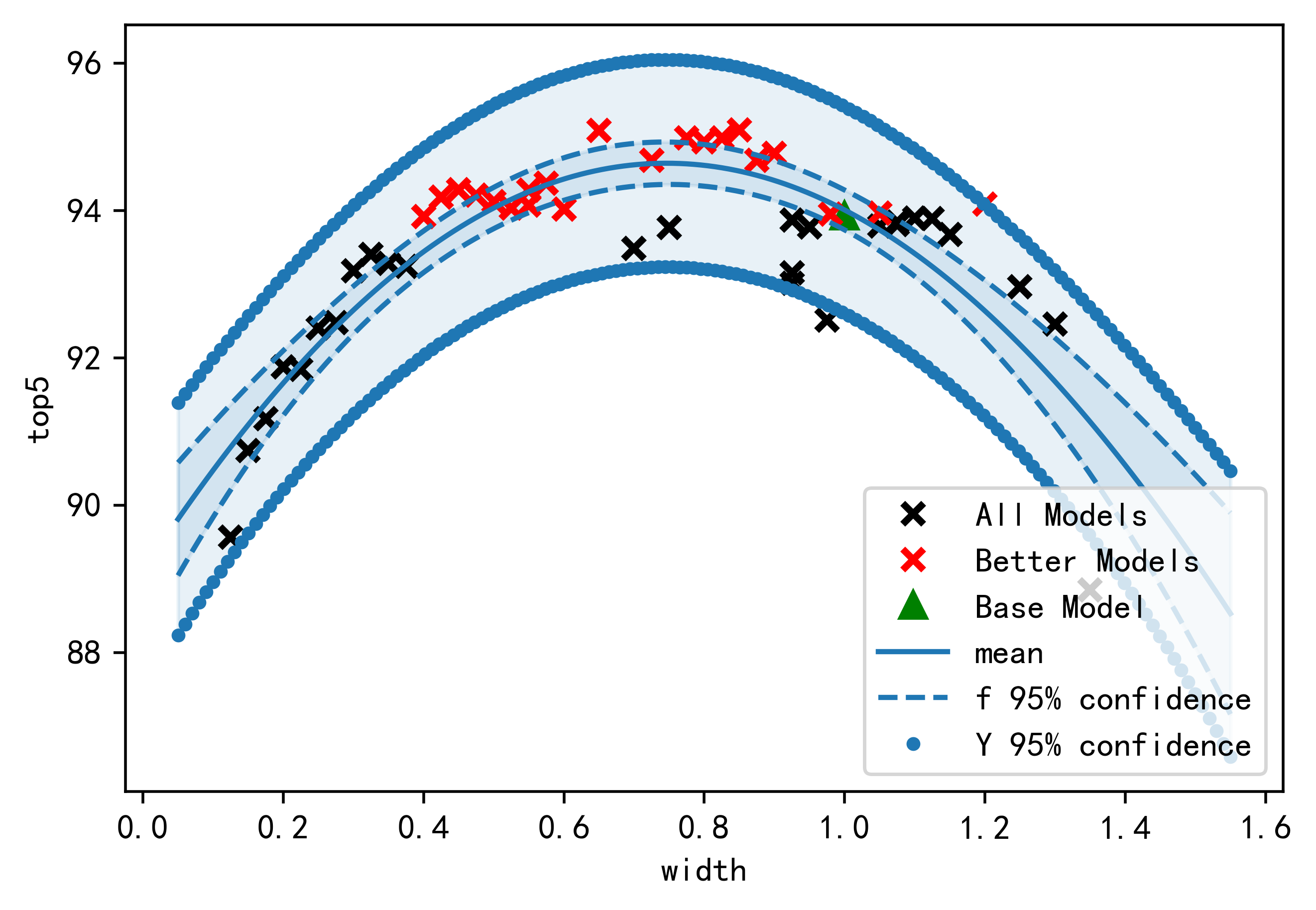} 
\\
& (c) Top 5 $\textit{v.s.}$ $d_m$ & (d) Top 5 $\textit{v.s.}$ $w$  
\end{tabular}
\caption{Fitted Top 5 accuracy $\textit{v.s.} (r, d_i,d_m, w)$ for MobileViT V2 models with $\sim$ 2000M MACS on ImageNet-100. This plot follows the similar labels used in top 1 accuracy. The performance peak is obtained in the median value of the factor and has a low in the small value of the factor. The overall variation is smooth, but not of exactly linear shape. }
\label{fig:top5-rdw}
\end{figure}

\textbf{Resolution and width are more significant:} We perform a series of experiments with different factors $(r, d_i, d_m, w)$ in MobileViT V2. 
To reach the MACs around $m_0$
, we randomly select the depth of inverted residual blocks, the depth of MobileViT blocks, and the width by tuning resolution to obtain different models. The resulting model has the target MACs.
Similarly, we do the same thing to tune width and depths.
To verify the performance, these models are trained with $100$ epochs on the ImaegNet-100 dataset. As shown in \fref{top1-rdw}, the accuracy is impacted by the resolution and width more, compared with the depths. When $d_i\geq1.5$ and $d_m\geq1.1$, the accuracy of MobileViT has reached a relatively stable state, with very few changes. Largely adjusting the depths has a relatively small impact on model performance.
We observe that the highest top 1 accuracy falls within the range of resolutions $1.0$ to $2.0$. For $r\leq1.0$, a larger resolution leads to higher accuracy, while the accuracy slightly decreases when $r$ exceeds $1.6$. 
As shown in subplot (a) \fref{top1-rdw}, the accuracy increased from about 62\% in $r=0.5$ to about 83\% in $r=1.5$. 
The width demonstrates a negative correlation with top 1 accuracy when $w\geq1.2$. Narrowly constraining the resolution and width may cause us to overlook some promising models. 
If we adhere to the MobileViT guidelines to achieve a model with $m_0$ MACs, namely MobileViT$^{m_{0}}$, where the four dimensions are computed and fine-tuned as $r=1.1$, $d_i=1.0$, $d_m=0.9$, $w=0.9$. However, its accuracy on ImageNet-100 is only $79.09\%$, falling short of the optimal combination for $m_0$ MACs. However, as illustrated in \fref{top1-rdw}, several MobileViTs with superior performance are randomly generated. This observation prompts us to explore a new formula for global architecture adjustment that can yield better MobileViTs within a fixed MACs constraint.


\subsection{Impact of joint factors}
In the design of efficient MobileViT, we must consider not only the impact of single factors but also the impact of joint factors. Therefore, to better understand the rule of MobileViT design, we explore how joint factors nonlinearly impact the performance and efficiency of MobileViT. 

\textbf{Modeling the impact of joint factors with a 2D grid GP:}
Mathematically, combinations of factors refer to choosing different sizes for two factors of MobileViT. 
Various methods exist for enumerating $2$-combinations, such as cross-validation. These methods are usually less efficient and unexplainable, which involves exhaustively enumerating all $2$ combinations of two factors. 
If both the sets of two factors have $n$ elements, the number of $2$-combinations is $n^{2}$. 
However, a 2D grid GP allows us to capture the joint impact of two factors, making it suitable for scenarios where joint factors nonlinearly determine the performance of MobileViT. The resulting 2D grid GP model not only fits the data exceptionally well but also offers insights into its properties through the interpretation of smoothness, variation, maximum, and minimum values on the posterior 2D grid.
For the fitting of 2D grid GP, we utilize all experimental observations as training samples, with every point on the grid serving as test data. 
Through our analysis, we find that the resolution and width, as a pair of joint factors, have a more significant impact. 

\begin{figure}[h!]
\centering
\renewcommand{\tabcolsep}{-0.5mm}
\scriptsize
\begin{tabular}{p{0.5mm}*{2}{c}}
& \figref{0.51}{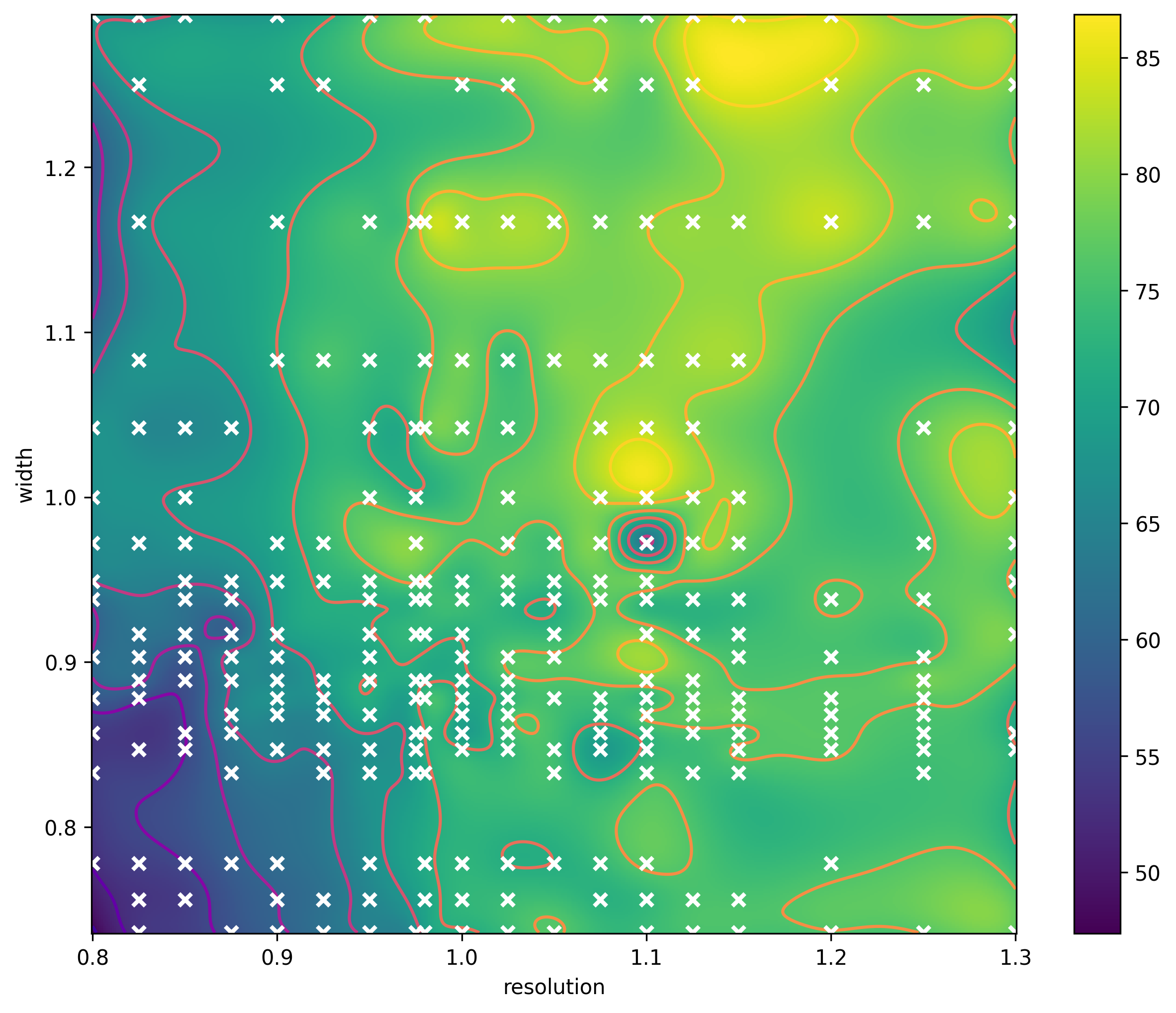} 
& \figref{0.51}{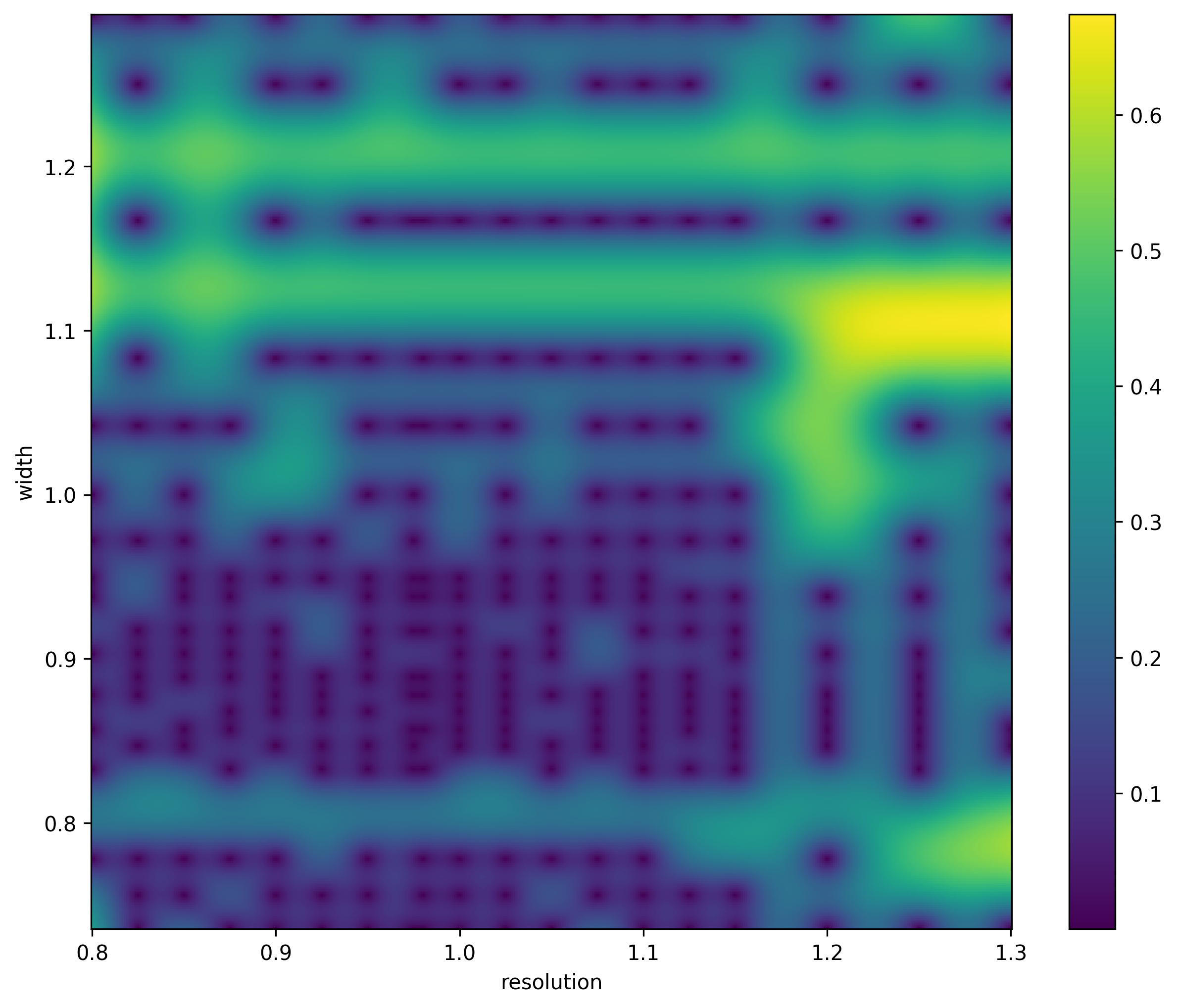} \\ 
\end{tabular}
\caption{Contour plot of the posterior mean (left) and variance (right) of the model performance predicted by the 2D grid GP fitting on joint factors resolution-width. Observations are labeled by white crosses. The color bar indicates the range of performance. 
}
\label{fig:rw-gp-mean}
\end{figure}    


\begin{figure}[h!]
\centering
\renewcommand{\tabcolsep}{-0.5mm}
\scriptsize
\begin{tabular}{p{0.5mm}*{2}{c}}
& \figref{0.51}{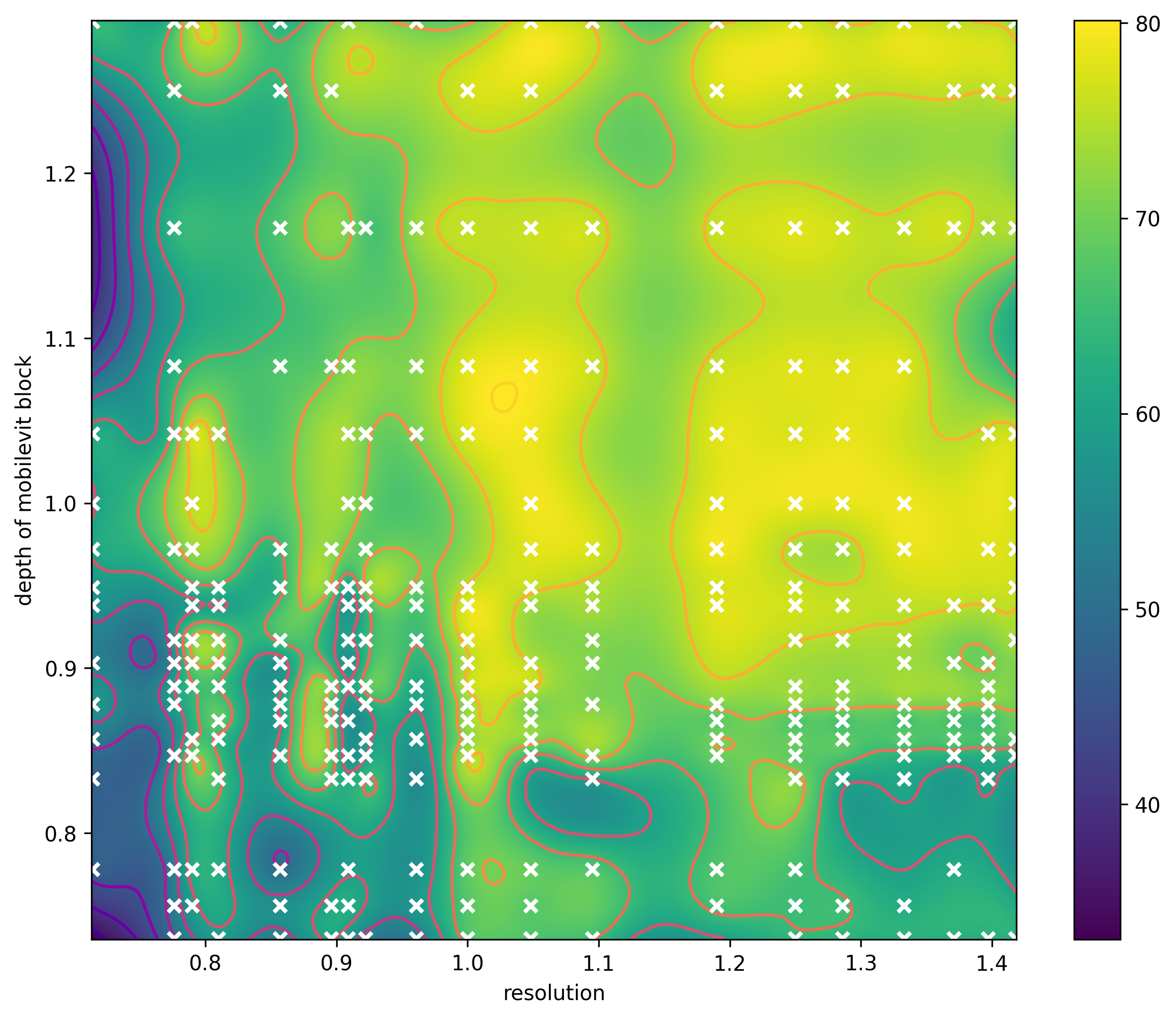} 
& \figref{0.51}{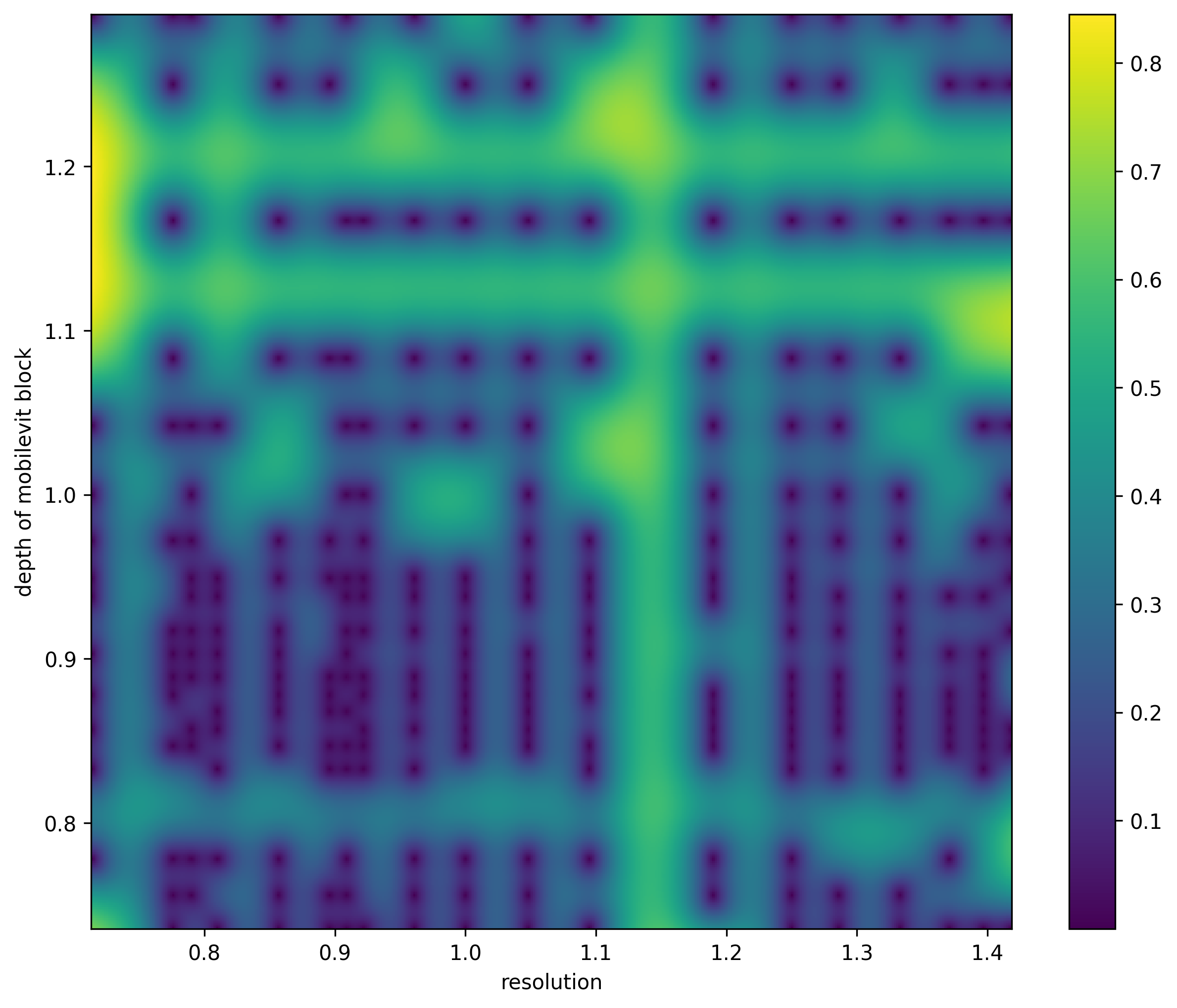} \\ 
\end{tabular}
\caption{Contour plot showing the performance of MobileViT in terms of posterior mean (left) and variance (right) as a function of the characteristic resolution and the depth of MobileViT block. Note, a ridge with $r=1.05$ and $d_m=1.08$. 
}
\label{fig:rdm-gp-mean}
\end{figure}   

\textbf{Joint impact of resolution and width:}
As shown in \fref{rw-gp-mean}, the joint factor of resolution and width both impact model performance significantly. 
This might be caused by the fact that high-resolution image contains more semantic information and a wider network extracts more semantic features. 
The training data lies on a regularly spaced grid. 
In this scenario, we model a 2D function with the training data situated on an evenly spaced grid within the range $(0.8, 1.3) \times (0.7, 1.3)$, containing $500$ grid points along each dimension. Hence, there exist $250,000$ combinations of joint factors that neither cross-validation nor NAS can efficiently explore. \fref{rw-gp-mean} illustrates the posterior mean and variance of the model performance as a function of the characteristic resolution and width.

From the posterior mean, we can see the 2D grid GP successfully reconstructs the model performances with different joint factors. The predictive model performance aligns with the training sample.
The latent function varies more smoothly. The model performance has a clear maximum around the factor values $\{r=1.1, w=1.0\}$. 
The model performance has a clear maximum around the factor values $\{r=1.1, w=1.0\}$. 
When $r$ values are less than $1.0$, a larger width slightly improves performance, while performance significantly increases when both $r$ and $w$ exceed $1.0$ and $1.0$, respectively.  As seen in the figure, uncertainty increases as we move away from the training samples. As would be expected, the nonlinear relationship is more complex for medium resolution and medium width.


\textbf{Joint impact of resolution and depth of MobileViT block:}
We also consider the joint impact of resolution and depth to investigate how model performance varies with their combinations. 
The posterior mean and variance of model performance are presented in \fref{rdm-gp-mean}, with our objective being to model the performance based on the joint factors resolution and depth of the MobileViT block. 
Additionally, note in \fref{rdm-gp-mean} that the 2D grid GP model produces relatively confident predictions.
In this case, the joint factors are distributed on a uniformly spaced grid within the range $(0.7, 1.45)\times(0.7, 1.3)$, containing $500$ grid points along each dimension. 
In \fref{rdm-gp-mean}, the highest top 1 accuracy is achieved at the location ${r=1.05, w=1.05}$ in the grid. For $r\leq1.0$,
the model performances are poor even given different depths. A larger depth slightly enhances performance, which confirms the impact of depth. 
These experiments demonstrate how resolution and width can promote non-trivial Mobilevit, and the GP's ability to examine single and joint factors from the data proves valuable in practice. Despite the absence of simple parametric assumptions, the GP's capacity to learn nonlinear underlying functions makes it an attractive model for navigating the efficiency of MobileViT.




\subsection{Efficiency rule for MobileViT}
Based on previous investigation, unlike factors of resolution and width,  MobileViT generally doesn't require many layers. Even with shallower layers MobileViT shows good performance.
Therefore, when designing the architecture of MobileViT, a relatively small number of layers in the MobileViT block is enough.

\begin{figure}[h!]
\centering
\renewcommand{\tabcolsep}{1.0mm}
\begin{tabular}{p{0.5mm}*{2}{c}}
& \figref{0.45}{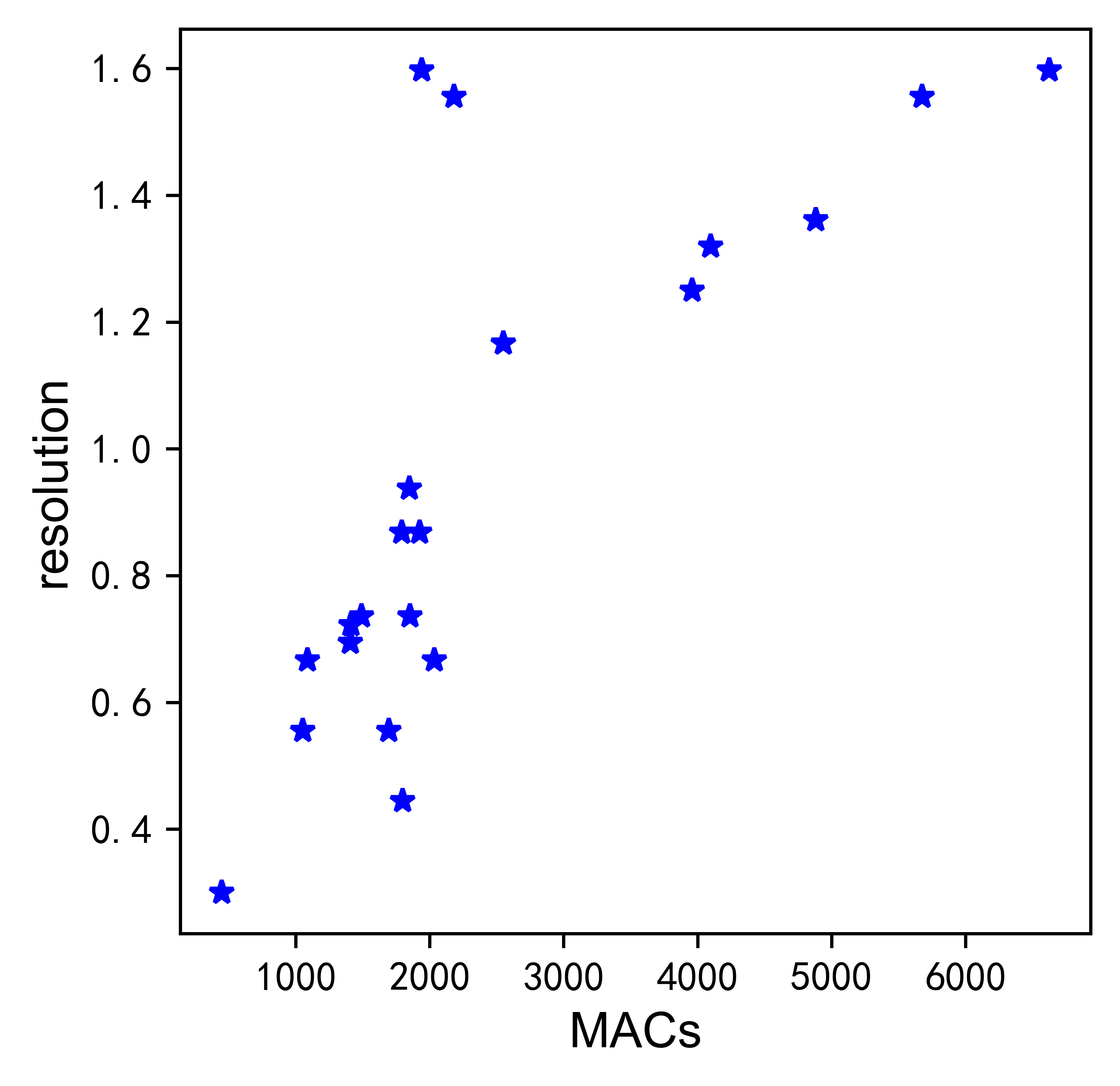} 
& \figref{0.45}{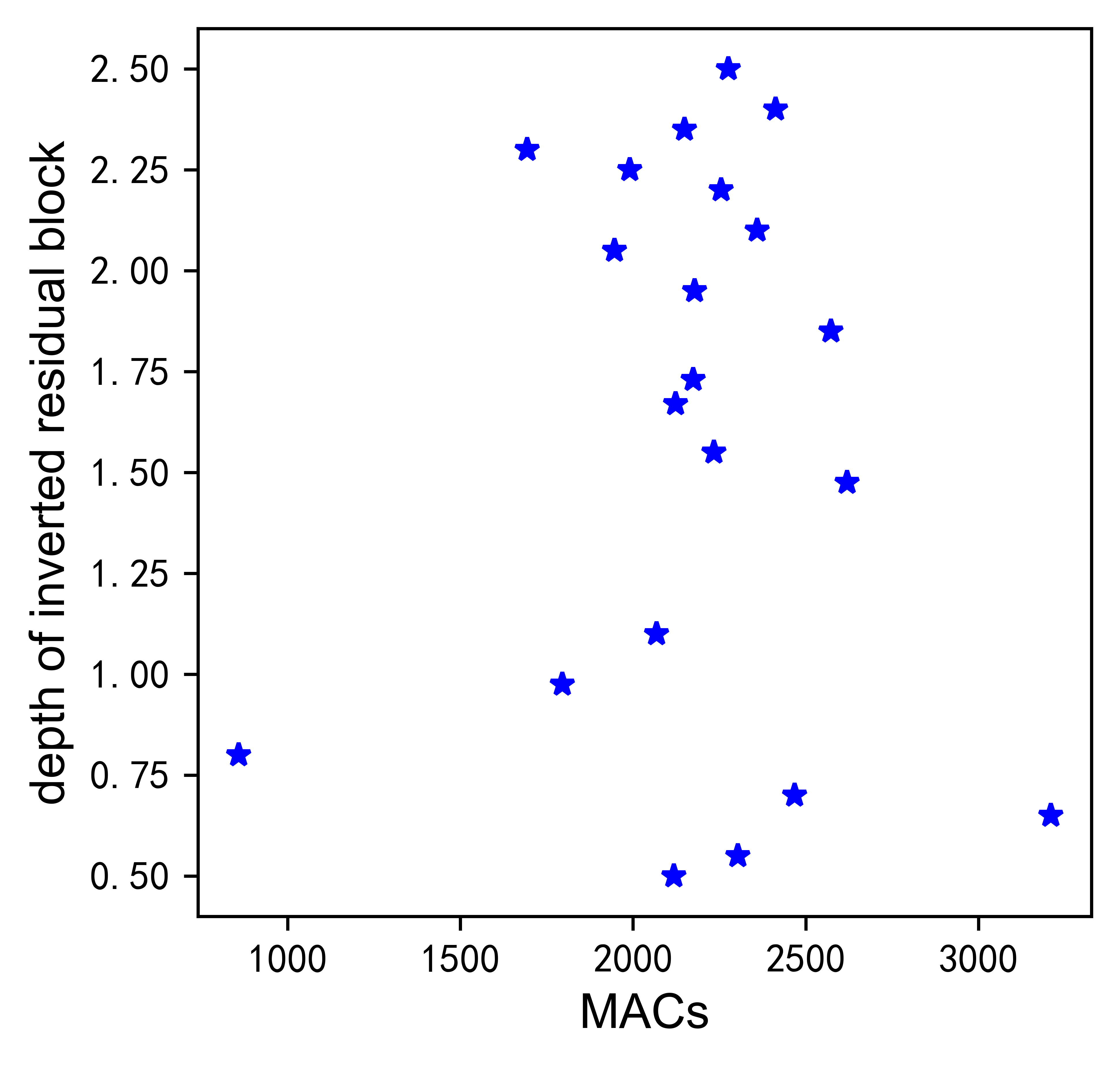} \\ 
& (a) $r$ & (b) $d_i$ \\ 
& \figref{0.45}{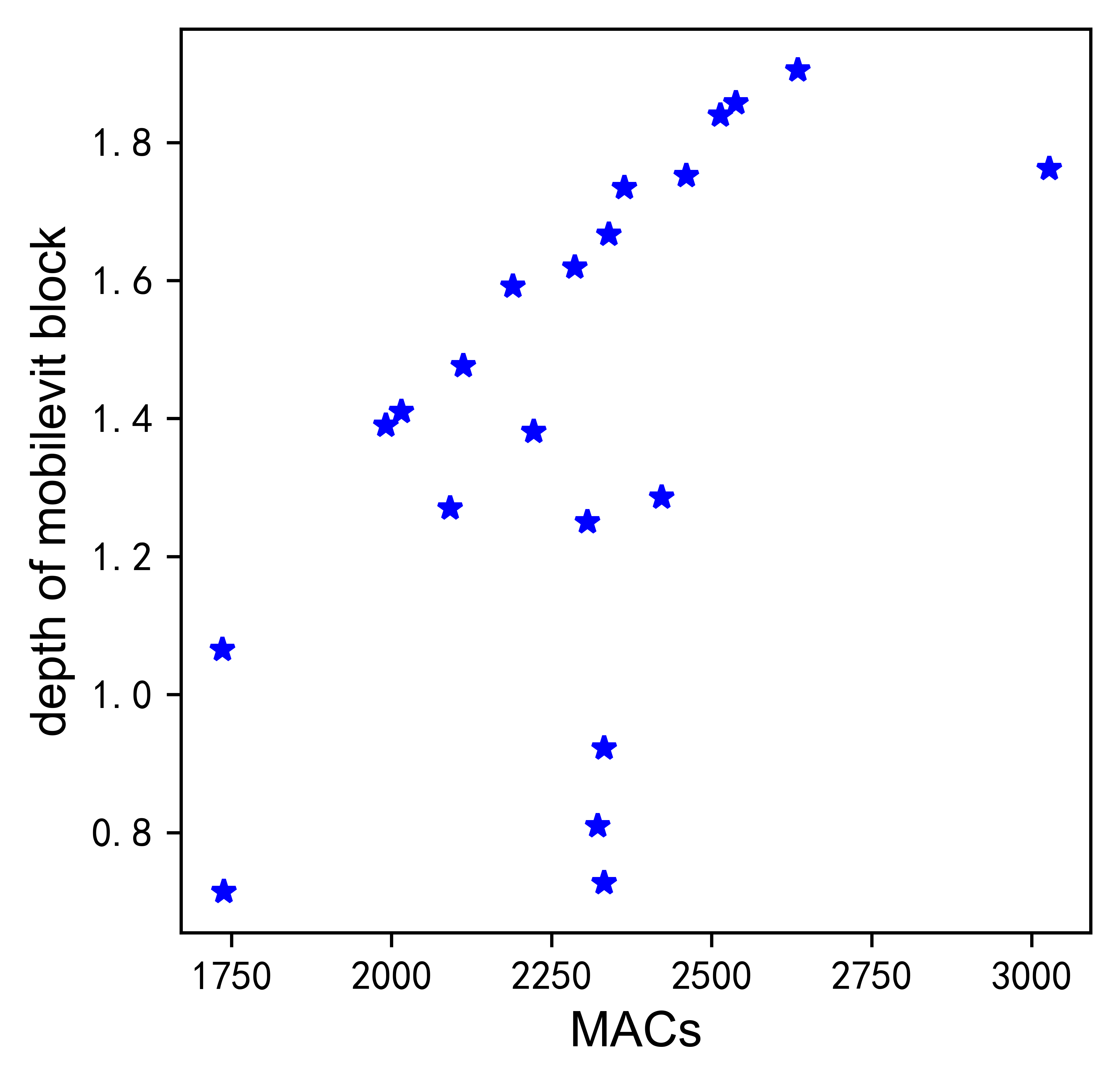}
& \figref{0.45}{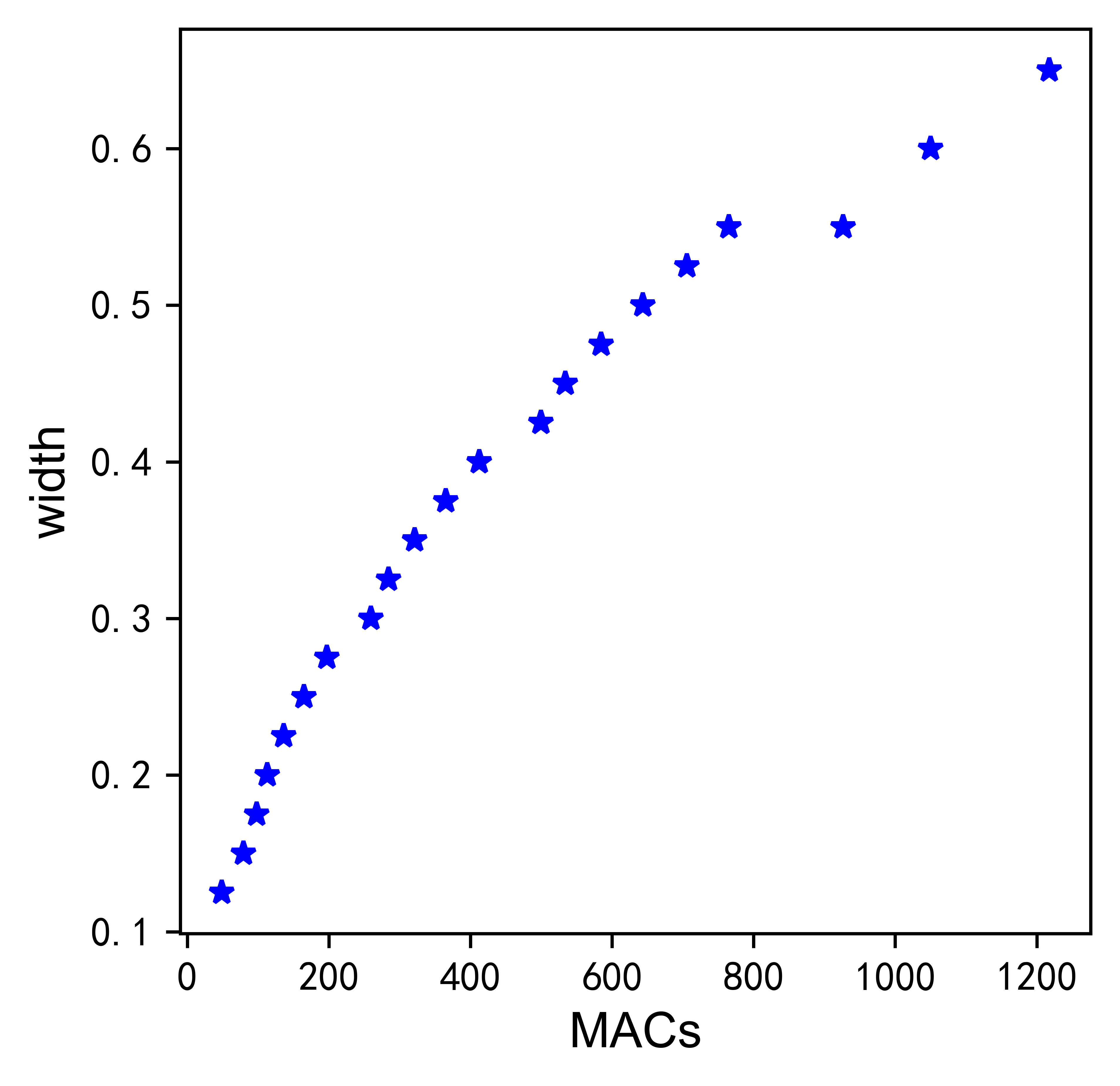} \\    
& (c) $d_m$ & (d) $w$ \\ 
\end{tabular}
\caption{$(r,d_i,d_m,w) \textit{v.s.}$ MACs for the models on Pareto front.}
\label{fig:pareto}
\end{figure}   

\textbf{Modeling efficiency rule as a GP:} Given a baseline MobileViT V2 model, we aim at finding a feasible way to shrink the over-parameterized model in four dimensions, $(r,d_i,d_m,w)$, with some constraints of MACs. 
To make our model exquisite, we introduce a reduction factor $c\in\mathbb{R}$ with $0<c<1$. Then, we assume that the optimal coefficients $r,d_i,d_m,w$ follow an underlying function of the reduction factor $c$ as
\begin{align}\label{eq:f}
r=f_r(c),\,\,d_i=f_{d_i}(c),\,\,d_m=f_{d_m}(c),\,\,w=f_w(c),
\end{align}
where $f_r(\cdot), f_{d_i}(\cdot), f_{d_m}(\cdot)$ and $f_w(\cdot)$ are functions describing the four factors. By verifying the models with different factors sampled randomly, we can learn the nonlinear relationship between $(r,d_i,d_m,w)$ and the model performance.

To delve deeper into the characteristics of the top-performing models, we focus on models situated along the Accuracy-MACs Pareto front. 
We set the constraint of Accuracy-MACs as follows.
\begin{align} 
\begin{cases}
    \min\:m, &  0.2< m <1.1m_0, \\
    \max\:acc, &  0.5< acc < 1,
\end{cases}  
\end{align}
where $m$ and $acc$ denote the MACs and accuracy of efficient MobileViT, respectively.
This front represents a collection of solutions that are non-dominated, considered optimal when no objective can be improved without compromising at least one other objective. Specifically, we also employ the NSGA-III nondominated sorting strategy \cite{deb2013evolutionary} to select the top 20\% of models exhibiting superior performance and lower computational complexities (i.e., MACs). The relationship between resolution, width, depths, and MACs for these chosen models is illustrated in \fref{pareto}. 
There is a stronger relationship between $r, d_i, d_m$ and MACs. 
Making use of the observed relationship between $(r, d_i, d_m, w)$ and MACs, we introduce a formula for adjusting the architecture of MobileViT. We compute the optimal combinations of $(r, d_i, d_m, w)$ and build models with high performance, while also satisfying the given requirement of MACs  $c\times m_0$. 
We recommend prioritizing adjustments of resolution and width to maintain the performance of lighter MobileViT.
To illustrate this point, we employ another GP \cite{williams2006gaussian,grunblatt2015determining,chen2021gaussian} to model the nonlinear relationship between  $c$ and architecture factors. Treating $\{(c_i, y_{i})\}_{i=1}^n$ as the training set with $n$ examples which are i.i.d, we have $y_{i}=f(c_i)+\varepsilon_i$, 
where $y_{i}$ is one of the four factors, $f(c_i)\sim\mathcal{GP}(0, k(c_i,c_i'))$,  
noise $\varepsilon_i\sim\mathcal{N}(0,\sigma_{n}^{2})$, and 
$k(c_i,c_i')$ is the kernel function. 

For the GP, 
we opt for a Mat\'ern kernel $k_{mat}$ to capture the desired correlation between pairs of observations.
Then we obtain the joint Gaussian distribution of the training set $C$ and the test point $c_{*}$:
\begin{align}\label{eq:gauss_R}
\begin{bmatrix}
\vy \\
y_{*}
\end{bmatrix}\sim\mathcal{N}\Big(
\mathbf{0},\begin{bmatrix}
    K + \sigma^2 I & \vk \\
   \vk^{\top} & k(c_{*}, c_{*}) + \sigma^2
\end{bmatrix}   
\Big),  
\end{align}
where $K=k_{mat}(C, C)$, $\vk=k_{mat}(C, c_{*})$. 
The posterior distribution $\mathcal{N}(\tilde{y}_{*}, \mathbb{V}[\tilde{y}]_{*})$ for the test point $c_{*}$ has an analytical form as 
\begin{align}
\label{eq:r_dist}
 \tilde{y}_{*}&=(K+\sigma^2I)^{-1}\vy, \\
\mathbb{V}[\tilde{y}]_{*} &= k(c_{*},c_{*}) + \sigma^2 - \vk^{\top} (K+\sigma^2I)^{-1}\vk.
\end{align}
Similarly, we can obtain the formula of depth $d_i$ by substituting $y$ with $d_i$. Given any network architecture, all formulas of architecture factors can simply follow this GP. Unlike the manually crafted compression method used in MobileViT, this shrinking rule is developed by observing superior small models. This approach can produce more efficient 
MobileViT with better performance. 

\section{Experiments}\label{sec:exp}
In this section, we employ our efficiency formula tailored for the magic 4D cube to downsize MobileViT. The efficacy of our approach is demonstrated through rigorous testing on prominent vision benchmarks. We consider diversified datasets and evaluation metrics for efficient models. Datasets used in this paper include ImageNet-100, TieredImageNet, Cifar-10, and Cifar-100. More details of the dataset can be found in the appendix.
We consider top 1 and top 5 accuracies as indicators of shrunk models by our rule. 


\textbf{Implementation details:} The MobileViT V2 and compressed MobileViT V2 are implemented using PyTorch and trained on NVIDIA GEFORCE RTX 4080. The batch size used in training is $64$. We employ the AdamW optimizer\cite{ilya2019decoupled} for the training of all models. For MobileViT V2 based models, we follow the same settings and configurations as \cite{mehta2022separable}. 

\subsection{Experiments on ImageNet-100} 
In this experiment, we employ random sampling to generate a variety of MobileViT models with distinct architecture factors.
Specifically, the resolution, depth, or width is randomly chosen within the 
range of $0.8\leq r \leq 1.7, 1.3\leq d_i \leq 2.3, 0.8\leq d_m \leq 1.6, 0.4\leq w \leq 1.2$. All models including the sampled MobileViT and MobileNetV1 undergo training on the ImageNet-100 dataset for 100 epochs, using the same training hyperparameters \cite{mehta2022separable}. 
\tref{performance_baseline} presents the results for all the models, revealing a general trend where the larger models obtain higher accuracy. 
The model of the baseline MobileViT V2 achieves 82.36\% accuracy. Our shrunken MobileViT V2 obtains top 1 accuracy with $83.21\%$ and top 1 accuracy with $95.73\%$.
Notably, it substantially outperforms other models. These findings underscore the need for a more effective efficiency rule shrinking MobileViT.
\begin{table}[hbt!]
\caption{Performances of MobileViT V2 on ImageNet-100.}
\label{tab:performance_baseline}
\renewcommand{\tabcolsep}{2.0mm}
\centering
\begin{tabular}{ccccc}
\toprule
Model           & Weights & Top 1 & Top 5   \\
\midrule

MobileNetV1     & 3.309 M    & 81.37\% &95.06\%   \\
MobileNetV2     & 2.352 M   &79.10\% &94.02\%    \\
MobileNetV3     & 4.285 M  &76.18\% &92.38\%\\
MobileViT V2    & 17.43 M  &82.36\% &95.33\% \\
MobileViT V2 (ours) & 13.69 M & 83.21\% &95.73\%\\
\bottomrule
\end{tabular}
\end{table}

\subsection{Experiments on TieredImageNet}
To further investigate the generalizability of our formula, we conducted additional experiments using the larger TieredImageNet dataset. As shown in
\tref{performance_tieredimagenet}, the best top 1 accuracy of MobileNet is 68.12\%. For the baseline MobileViT V2 model situated along the Accuracy-MACs Pareto front, it achieves top 1 accuracy with 69.50\%. However, due to the use of the proposed efficiency rule, our proposed method consistently outperforms both the baseline MobileViT V2 and MobileNet. 
\begin{table}[hbt!]
\caption{Performances of MobileViT V2 on TieredImageNet}
\label{tab:performance_tieredimagenet}
\renewcommand{\tabcolsep}{1.5mm}
\centering
\begin{tabular}{cccc}
\toprule
Model      & Weights& Top 1 &Top 5   \\
\midrule


MobileNetV1&3.55 M &68.12\% &88.82\%\\
MobileNetV2&2.66 M &63.38\%&86.21\%\\
MobileNetV3&4.80 M &63.69\% &86.39\%\\
MobileViT V2 & 19.51 M & 69.50\%&89.84\%\\
MobileViT V2 (ours)  & 17.43M & 66.73\% & 87.38\% \\
\bottomrule
\end{tabular}
\end{table}

\subsection{Experiments on Cifar-10 and Cifar-100}
In \tref{performance-cifar10} and \tref{performance-cifar100}, 
we present the results of experiments on Cifar-10 and Cifar-100 for all the models. These experiments reveal that our shrunken MobileViT V2 still outperforms other models. Our shrunken MobileViT V2 surpasses the baselines in both datasets. 
As shown in \tref{performance-cifar10} and \tref{performance-cifar100}, the overall top 1 and top 5 accuracies indicate a performance gain of our MobileViT V2. Both top 1 and top 5 demonstrate, unsurprisingly, that MobileViT using the rule of efficiency is superior to the original MobileViT. Notably, compared to the competitive baseline, Our shrunken MobileViT V2 outperforms it by 0.8\%.
These findings emphasize the necessity for a more efficient rule to enhance the compactness of MobileViT.

\begin{table}[hbt!]
\caption{Performances of MobileViT V2 on Cifar-10}
\label{tab:performance-cifar10}
\renewcommand{\tabcolsep}{2.0mm}
\centering
\begin{tabular}{cccc}
\toprule
Model            & Weights &  top 1   & top 5   \\
\midrule

MobileNetV1      & 3.56 M    &  78.25\% & 96.81\%       \\
MobileNetV2       & 2.24 M    &  88.32\% & 99.62\% \\
MobileNetV3        & 4.17 M    & 88.33\% & 99.58\% \\
MobileViT V2 & 17.43M & 90.15\% &99.35\%\\
MobileViT V2 (ours)  & 13.73 M   & 90.51\%&99.58\% \\
\bottomrule
\end{tabular}
\end{table}

\begin{table}[hbt!]
\caption{Performances of MobileViT V2 on Cifar-100}
\label{tab:performance-cifar100}
\renewcommand{\tabcolsep}{2.0mm}
\centering
\begin{tabular}{cccc}
\toprule
Model      & Weights & Top 1   & Top 5   \\
\midrule

 MobileNetV1      & 3.31 M    & 70.16\%  &90.86\%      \\
 MobileNetV2      & 2.35 M    & 67.34\%  & 89.88\%        \\
MobileNetV3      &  4.28 M    & 68.94\%  & 90.10\%     \\
MobileViT V2 &17.43 M &67.38\%&88.79\%\\
MobileViT V2 (ours)  & 15.27 M  & 67.98\% & 89.21\%\\
\bottomrule
\end{tabular}
\end{table}

\subsection{Experiments on CUB\_200\_2011 dataset and Car Parts dataset}

We have augmented our study with new experiments on expanded datasets including CUB\_200\_2011, and Cars Parts in \tref{performance-cub} and \tref{performance-cars-parts}. These experiments provide a multifaceted perspective on the value of our approach.
In addition, our focus is on a method to optimize the architectures of existing models rather than proposing a new model architecture. We believe that our extended experiments are sufficient to capture the complexity and variability required to thoroughly assess the optimized model's performance across different scenarios.

\begin{table}[hbt!]
\caption{Performances of MobileViT V2 on CUB\_200\_2011 }
\label{tab:performance-cub}
\renewcommand{\tabcolsep}{2.0mm}
\centering
\begin{tabular}{cccc}
\toprule
Model            & top1 &  top 5   & MACs   \\
\midrule
EfficientFormer V1\cite{li2022efficientformer}  & 59.73 & 79.81 & 1.9G  \\
 EfficientFormer V2\cite{li2023rethinking}  & 60.74 & 83.42 & 1.9G \\
 EdgeViT \cite{pan2022edgevits}            & 61.52 & 85.26 & 1.9G \\
 FastViT\cite{vasu2023fastvit}           & 56.25 & 76.56 & 1.9G \\
 MobileViT V2        & 63.83 & 87.81 & 1.9G \\
 MobileViT V2 (ours) & 64.43 & 88.81 & 1.9G\\ 
\bottomrule
\end{tabular}
\end{table}

\begin{table}[hbt!]
\caption{Performances of MobileViT V2 on Car Parts dataset}
\label{tab:performance-cars-parts}
\renewcommand{\tabcolsep}{2.0mm}
\centering
\begin{tabular}{cccc}
\toprule
Model            & top1 &  top 5   & MACs   \\
\midrule
EfficientFormer V1  &87.83 & 97.30 & 1.9G  \\
 EfficientFormer V2  & 87.43 & 96.97 & 1.9G  \\
 EdgeViT             & 84.50 & 95.00 & 1.9G \\
 FastViT             & 86.71 & 96.87 & 1.9G \\
 MobileViT V2        &  89.87 & 98.38 & 1.9G  \\
 MobileViT V2 (ours) & 92.34 & 98.90 & 1.9G \\
\bottomrule
\end{tabular}
\end{table}

\subsection{Latency on iPhone12}

We have measured the latency of MobileViT on a mobile device  (iPhone 12).  We have reported the latency of the model in \tref{performance-laten}. This table reveal that our optimized model has not only achieved a marked improvement in performance but also reduced inference time, further validating the effectiveness of our approach. While MobileViT exhibits a higher inference time compared to other ViT models, its performance is significantly enhanced.

Due to the universality and applicability of our approach, it can also inspire subsequent research to further optimize the structures of other ViT models, reduce inference time, and enhance performance. Through these experiments, researchers and developers can gain insights into the performance of MobileViT models on mobile devices and make informed decisions about their deployment and optimization.

\begin{table}[hbt!]
\caption{Latency of different models on iPhone12}
\label{tab:performance-laten}
\renewcommand{\tabcolsep}{2.0mm}
\centering
\begin{tabular}{cc}
\toprule
Models            & Latency   \\
\midrule
EfficientFormer V1 & 3.37 ms \\
EfficientFormer V2 & 2.96 ms \\
EdgeViT            & 3.41 ms \\
FastViT            & 3.22 ms \\
MobileViT V2       & 4.58 ms \\
MobileViT V2 (ours) & 4.27 ms \\
\bottomrule
\end{tabular}
\end{table}

\subsection{Visualization of learning curves}
To effectively showcase the impact of our approach, we illustrate the learning curves of our efficient MobileViT V2 in \fref{learn-curve}. The learning curves in \fref{learn-curve} include both validation loss and validation top 1. As depicted in \fref{learn-curve}, our MobileViT V2 maintains a significant convergence rate throughout the validation process. With EMA, the learning curves become smoother, which leads to a fast convergence of the training.
We randomly choose several models, each adjusted with our proposed four factors, to showcase the validation loss. In \fref{learn-curve}, the average loss along with the upper and lower bounds of the selected models. These bounds are determined as the average plus 1.96 times the standard deviation and the average minus 1.96 times the standard deviation, respectively.



\begin{figure}[htb!]
\centering
\renewcommand{\tabcolsep}{-0.1mm}
\scriptsize
\begin{tabular}{p{0.5mm}*{2}{c}}
& \figref{0.50}{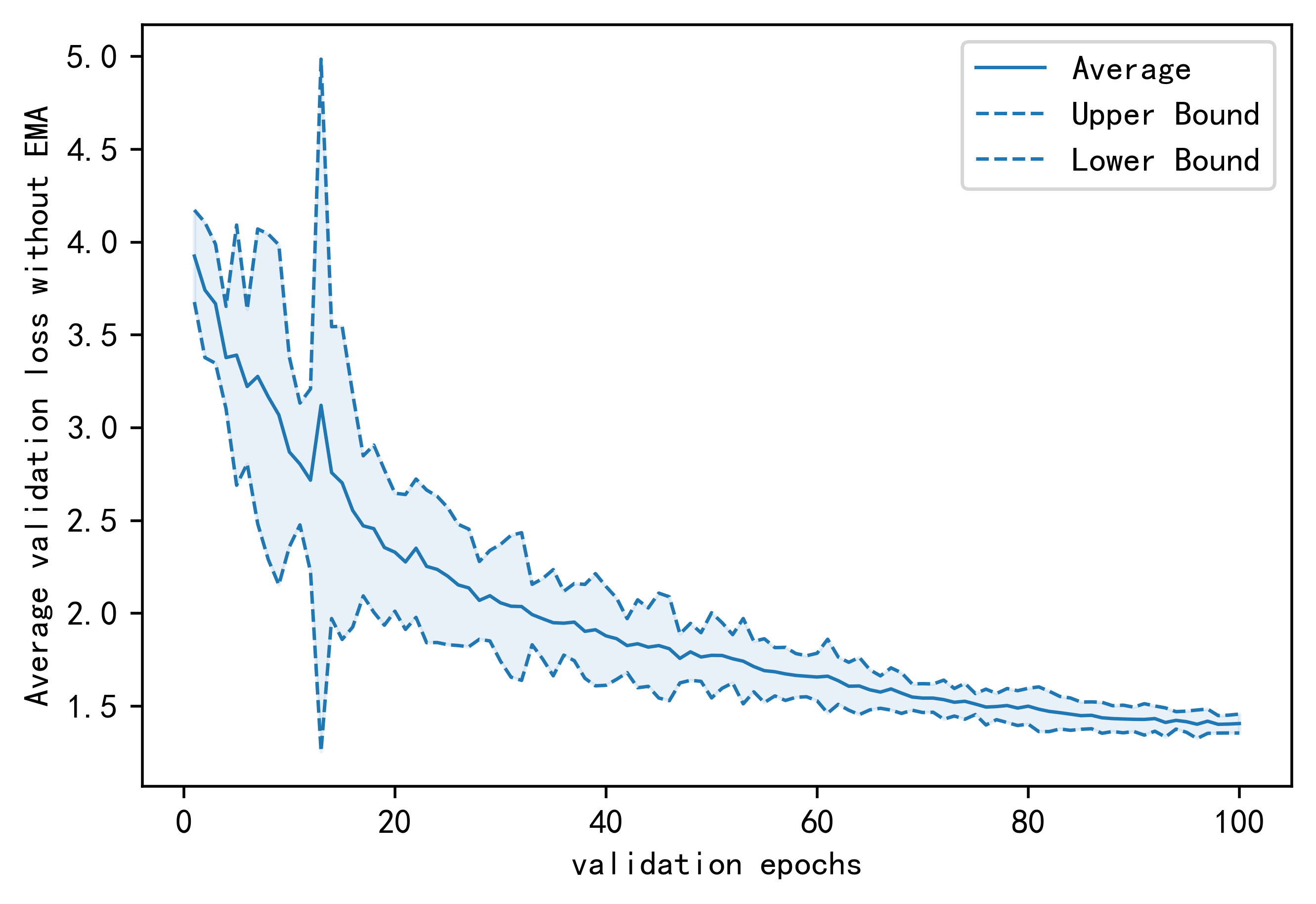} 
&  \figref{0.5}{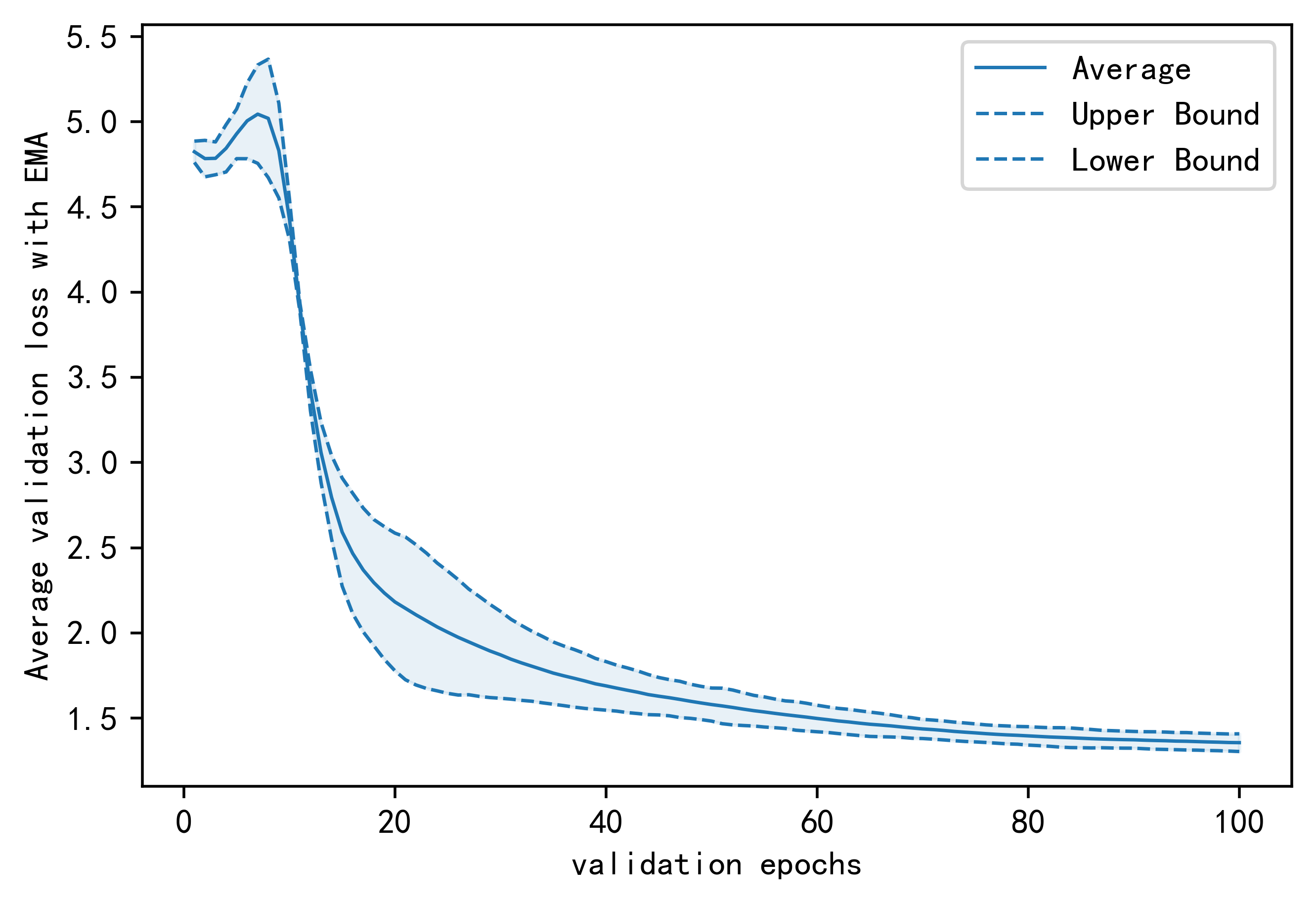} \\ 
& (a) Validation loss w/o EMA & (b) Validation loss w/ EMA \\ 
& \figref{0.5}{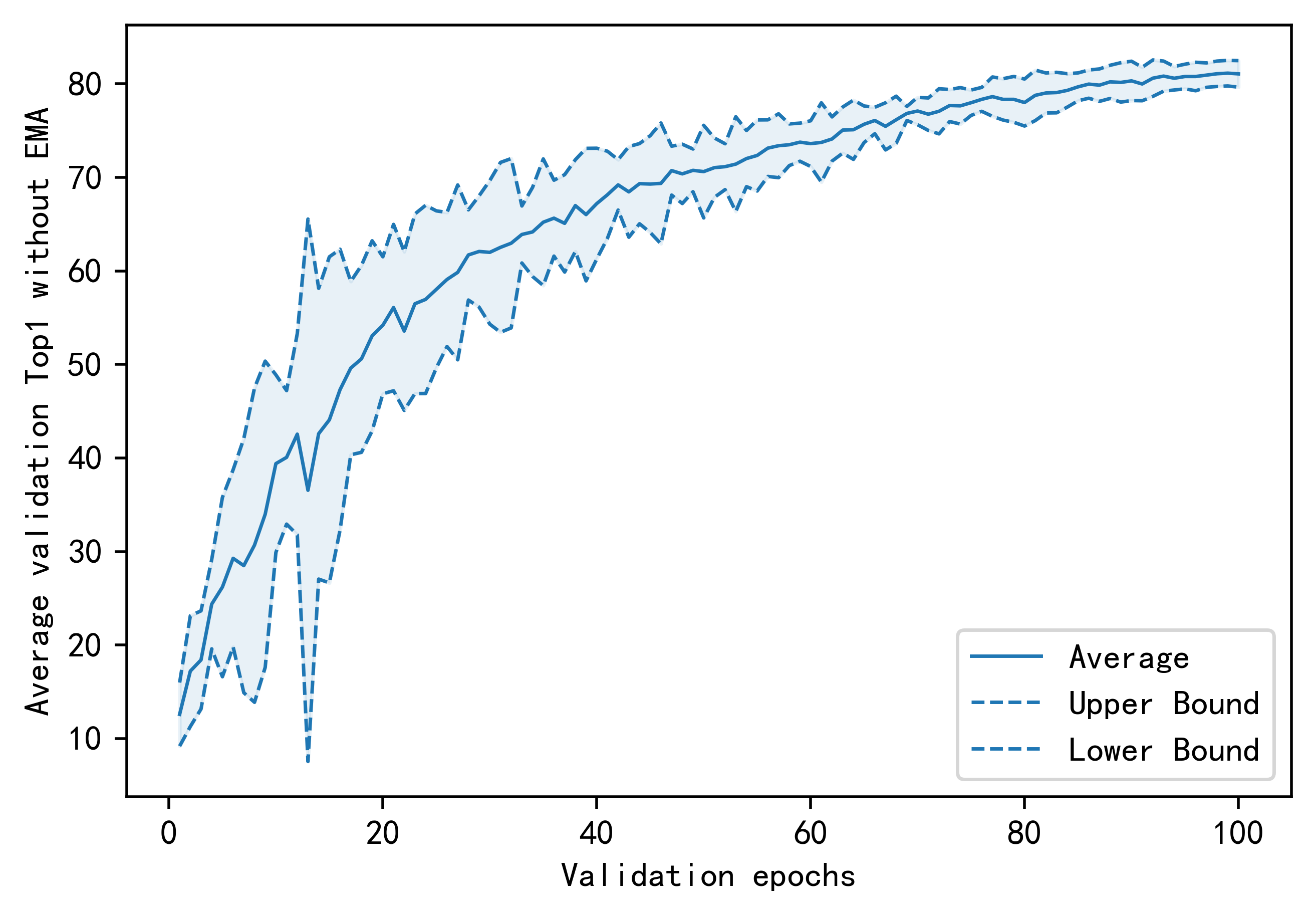}
& \figref{0.5}{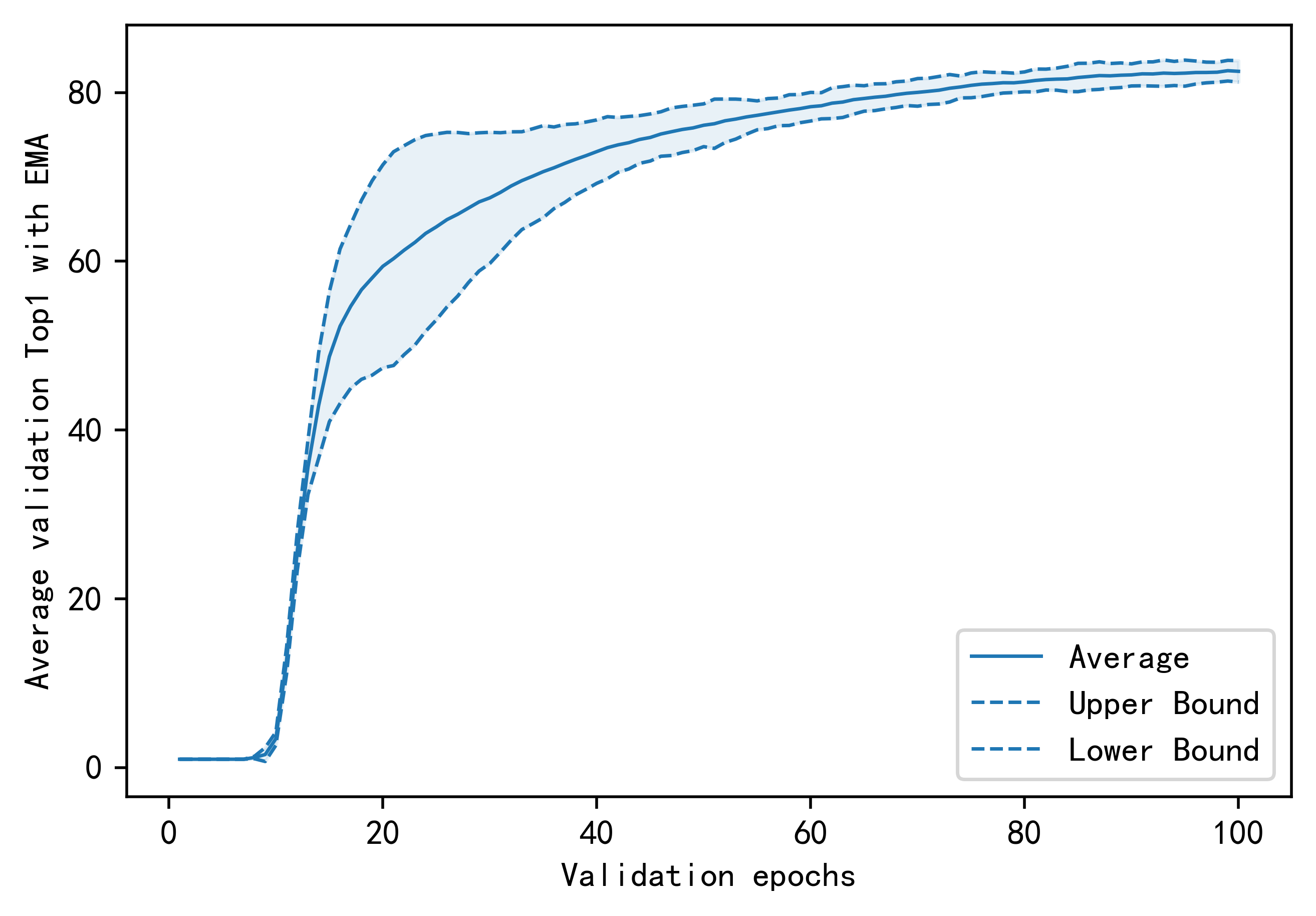} \\
& (c) Validation top 1 w/o EMA & (d) Validation top 1 w/ EMA  \\
\end{tabular}
\caption{Validation loss and validation top 1 accuracy with 100 epochs for both scenarios: without (w/o) and with (w/) EMA. This learning curve is obtained from the training on ImageNet-100}
\label{fig:learn-curve}
\end{figure}


\section{Conclusion}\label{sec:conclude}
In this paper, we systematically explore the efficiency-centric MobileViT architecture using GP to understand the relationship between performance and key architecture factors. The proposed downsizing formula proves highly effective. 
We consider twisting 4D architecture factors including resolution, depth of the inverted residual block, depth of MobileViT block, and width to gain a smaller and better model.
In particular, our proposed formula efficiently minimizes the search space of architecture factors.  
focusing solely on classification experiments limits the breadth of our study. 
Constraints imposed by device limitations hinder a full exploration of our approach's potential capabilities. A broader investigation into tasks such as object detection and semantic segmentation would provide a more comprehensive assessment of our method. Enhanced computational resources in future research can lead to a more comprehensive evaluation across diverse applications and domains, potentially yielding further improvements. \\

\bibliographystyle{ieeetr}
\bibliography{reference.bib}

\newpage
\appendix
\onecolumn

\end{document}